\documentclass[preprint,12pt]{elsarticle} % authoryear

\usepackage{amssymb}

\usepackage{url}            % simple URL typesetting
\usepackage{booktabs}       % professional-quality tables

\usepackage{bm}

\usepackage{nicefrac}       % compact symbols for 1/2, etc.
\usepackage{microtype}      % microtypography
\usepackage{adjustbox}
\usepackage[flushleft]{threeparttable}
\usepackage{tcolorbox}
\usepackage[page, header]{appendix}

\usepackage{enumerate}

\newcommand{\EX}{\mathbf{E}}
\newcommand\kgiven[1][]{\:#1\vert\:}
\newcommand{\piprime}{\pi^{\prime}}
\newcommand{\pitheta}{\pi_{\theta}}
\newcommand{\pithetazero}{\pi_{\theta_0}}
\newcommand{\OPT}{\mathcal{O}}

\newcommand{\LEPSILONESTAR}{L_{\epsilon^\ast}}

\usepackage{bbm}

\usepackage{adjustbox}

\usepackage{enumitem}
\usepackage{mathtools}

\usepackage{subcaption}
\usepackage[utf8]{inputenc} % allow utf-8 input
\usepackage[T1]{fontenc}    % use 8-bit T1 fonts
\usepackage{url}            % simple URL typesetting
\usepackage{booktabs}       % professional-quality tables
\usepackage{nicefrac}       % compact symbols for 1/2, etc.
\usepackage{microtype}      % microtypography
\usepackage{xcolor}
\usepackage{graphicx}
\usepackage{booktabs} % for professional tables
\usepackage{float}
\usepackage{algorithmic}
\usepackage{algorithm}
\usepackage{stackengine}
\usepackage{listings}

\usepackage{comment}
\usepackage{apxproof}
\usepackage{thmtools, thm-restate}
\usepackage{wrapfig}
\declaretheorem{theorem}

\declaretheorem{lemma}
\declaretheorem{corollary}

\declaretheorem{definition}

\DeclareMathAlphabet\mathbfcal{OMS}{cmsy}{b}{n}
\usepackage{bm}


\usepackage{soul}
\usepackage{witharrows}

\definecolor{codegreen}{rgb}{0,0.6,0}
\definecolor{codegray}{rgb}{0.5,0.5,0.5}
\definecolor{codepurple}{rgb}{0.58,0,0.82}
\definecolor{backcolour}{rgb}{0.95,0.95,0.92}

\lstdefinestyle{mystyle}{
    backgroundcolor=\color{backcolour},   
    commentstyle=\color{codegreen},
    keywordstyle=\color{magenta},
    numberstyle=\tiny\color{codegray},
    stringstyle=\color{codepurple},
    basicstyle=\ttfamily\footnotesize,
    breakatwhitespace=false,         
    breaklines=true,                 
    captionpos=b,                    
    keepspaces=true,                 
    numbers=left,                    
    numbersep=5pt,                  
    showspaces=false,                
    showstringspaces=false,
    showtabs=false,                  
    tabsize=2
}

\usepackage{titletoc}
\usepackage{lipsum}

\usepackage[hidelinks, linkcolor=black, 
            anchorcolor=black, 
            citecolor=black,       
            ]{hyperref}

\usepackage{amsmath,amsfonts,bm}

\def\eqref#1{equation~\ref{#1}}
\def\1{\bm{1}}

\DeclareMathAlphabet{\mathsfit}{\encodingdefault}{\sfdefault}{m}{sl}
\SetMathAlphabet{\mathsfit}{bold}{\encodingdefault}{\sfdefault}{bx}{n}

\DeclareMathOperator*{\argmax}{arg\,max}

\usepackage{microtype}
\usepackage{graphicx}

\usepackage{booktabs} % for professional tables

\usepackage{hyperref}
\usepackage{amssymb}
\usepackage{amsmath}
\usepackage{amsthm}

\usepackage{adjustbox}

\usepackage{cancel}

\usepackage{bm}
\usepackage{makecell}
\usepackage[figuresright]{rotating}

\usepackage{adjustbox}

\usepackage{enumitem}

\usepackage{graphicx}
\usepackage{array}

\usepackage{extarrows}
\usepackage{caption}
\usepackage{graphicx}
\usepackage{sidecap}
\usepackage{float}
\usepackage{framed}
\usepackage{tabularx}
\usepackage{twoopt}
\usepackage{adjustbox}

\usepackage{lscape}

\usepackage{tikz}
\usetikzlibrary{fit}

\usepackage{booktabs,threeparttable}

\usepackage{lineno}

\newcommandtwoopt\Textbox[5][7.2cm][2cm]{%
\begin{tikzpicture}[remember picture,overlay]
  \coordinate (aux) at ([xshift=#1]#4);
  \node[inner ysep=3pt,yshift=1ex,draw=pink,thick,
    fit=(#3) (aux),baseline] 
    (box) {};
  \node[text width=#2,anchor=north east,
    font=\sffamily\footnotesize,
  align=right
    ] 
    at (box.north east) {#5};
\end{tikzpicture}%
}

\hypersetup{
colorlinks=true,
citecolor=red,
    linkcolor=red,
    filecolor=magenta,      
    urlcolor=red,
linktocpage}

\usepackage{lipsum}


\begin{document}

\begin{frontmatter}
\title{SCPO: Safe Reinforcement Learning with Safety Critic Policy Optimization}

% A Survey of Human-Centered Safe Robot Reinforcement Learning Methods 

% % to Make Decisions

% Human-centered safe learning for autonomous agents

% % Authors and affiliations
\author{Jaafar Mhamed\fnref{label1}%
}
\author{Shangding Gu\fnref{label1}$^{*}$%
}

\cortext[cor1]{Correspondence to: \textit{{shangding.gu@tum.de}} 
\fntext[label1]{Technical University of Munich,  Munich, 85748, Germany.%
}
% \fntext[label2]{Technical University of Darmstadt,  Darmstadt, 64289, Germany. %
% }
% \fntext[label3]{King’s College London, London, WC1E 6EB, UK.%
% }
% \fntext[label4]{Tongji University,  Shanghai, 201804, China.%
% 
}

% \end{frontmatter}

\begin{abstract}
Incorporating safety is an essential prerequisite for broadening the practical applications of reinforcement learning in real-world scenarios. To tackle this challenge, Constrained Markov Decision Processes (CMDPs) are leveraged, which introduce a distinct cost function representing safety violations. In CMDPs' settings, Lagrangian relaxation technique has been employed in previous algorithms to convert constrained optimization problems into unconstrained dual problems. However, these algorithms may inaccurately predict unsafe behavior, resulting in instability while learning the Lagrange multiplier. This study introduces a novel safe reinforcement learning algorithm, Safety Critic Policy Optimization (SCPO). In this study, we define the safety critic, a mechanism that nullifies rewards obtained through violating safety constraints. Furthermore, our theoretical analysis indicates that the proposed algorithm can automatically balance the trade-off between adhering to safety constraints and maximizing rewards. The effectiveness of the SCPO algorithm is empirically validated by benchmarking it against strong baselines. 

\end{abstract}


% \begin{highlights}
% \item A review of safe Reinforcement Learning (RL) methods is provided with theoretical analysis and  application analysis.
% \item The key question that safe RL needs to answer is proposed, and five problems \textbf{``2H3W"} are analysed to address the key question.
% \item To examine the effectiveness of safe RL methods, several safe single-agent and multi-agent RL benchmarks are investigated.  
% \item The challenging problems are pointed out to guide the research directions.
% \end{highlights}

% \textbf{``2H3W"} problems;

\end{frontmatter}

\clearpage
\tableofcontents
\clearpage

\section{Introduction}
\label{chapter:introduction}

% The field of reinforcement learning (RL) involves an agent learning to take actions in an environment to maximize a long-term reward signal \cite{RichardS.Sutton}. It has been widely adopted in several fields such as finance \cite{rl_finance, rl_finance_2}, robotics~\cite{abeyruwan2023sim2real, gu2023safe}, transportation schedule \cite{rl_transport_2, rl_transport} and autonomous driving \cite{kiran2021deep,  rl_autonomous_2, rl_autonomous_3, gu2022constrained}.
% However, the absence of safety assurance significantly hinders the practical application of RL algorithms to real-world problems~\cite{gu2023safe, gu2023human}. RL agents rely on reward signals to make decisions, which can disregard safety restrictions. 

The field of reinforcement learning (RL) is centered on enabling an agent to learn actions in an environment to maximize a long-term reward signal \cite{RichardS.Sutton}. This approach has found widespread adoption across various domains, including finance \cite{rl2010finance, rl2003finance2}, robotics \cite{abeyruwan2023sim2real, gu2023safe}, transportation scheduling \cite{rltransport2, rl2020transport}, and autonomous driving \cite{gu2022constrained, kiran2021deep, rl2021autonomous, rl2018autonomous3}. Nevertheless, the lack of safety assurance poses a significant obstacle to the practical implementation of RL algorithms in real-world situations \cite{gu2023human, gu2023safe}. RL agents predominantly depend on reward signals for decision-making, which can inadvertently lead to the overlooking of crucial safety constraints. For example, an RL agent responsible for controlling a self-driving vehicle might receive a significant reward for high-speed driving; however, this behavior could raise the risk of collisions with other objects. RL agents may prioritize maximizing rewards over maintaining safe behavior, potentially leading to hazardous or catastrophic outcomes \cite{saferlimportance}.

In the context of this field, a primary methodology is the pursuit of a safe policy for ensuring the safety of RL applications. Notably, these methods~\cite{cpo, gu2023safe, Ray} propose the segregation of safety specifications from task performance as a pivotal approach. This conceptual demarcation serves as a catalyst for the introduction of a dedicated cost function, distinct from the conventional reward function. Consequently, this novel formulation gives rise to what is referred to as a Constrained Markov Decision Process (CMDP). The inclusion of constraint components in CMDPs augments the flexibility in modeling problems that entail trajectory-based constraints, offering an approach that ensures RL applications' safety by addressing constraints. However, extant algorithms based on CMDPs are beset by the limitation of low sample efficiency \cite{lowsampleefficiency}, which results in protracted convergence times and heightened exposure to unsafe behaviors during training. A majority of these approaches resort to employing a Lagrange multiplier to transmute the safety constraint problem into an unconstrained equivalent.

% The added constraint component in CMDPs enhances the flexibility of modeling problems with trajectory-based constraints, as opposed to alternative approaches that adapt immediate costs in MDPs to address constraints. Existing algorithms are hindered by low sample efficiency \cite{lowsampleefficiency}, leading to slow convergence and increased exposure to unsafe behaviors during training. The majority of approaches employ a Lagrange multiplier to transform the safety constraint problem into an unconstrained equivalent.

% model-free, on-policy

% To address the above issues, we propose a new algorithm, safety critic policy optimization (SCPO). Improving the return mostly contradicts respecting safety constraints. Therefore, balancing return and cost during training is not trivial. The core idea behind our method is to nullify the reward obtained from visiting unsafe states. We use a safety critic to approximate the safety of a state action pair and reduce its reward accordingly. The safety critic can be initialized pessimistically; every state is unsafe until proven otherwise. This leads to safer policies throughout training.
% SCPO aligns both objectives of maximizing return and adhering to safety constraints which leads to greater sample efficiency; training time is shorter, and unsafe trajectories are rarely generated.

To address the aforementioned challenges, we introduce a novel algorithm called Safety Critic Policy Optimization (SCPO). Striking a balance between improving the return and adhering to safety constraints during training is a non-trivial task, as these objectives often contradict each other. The central concept of our method is to nullify the reward obtained from visiting unsafe states. To achieve this, we employ a safety critic to approximate the safety of a state-action pair and adjust its reward accordingly. The safety critic can be initialized pessimistically, deeming every state as unsafe until proven otherwise, resulting in safer policies throughout the training process. SCPO effectively aligns both objectives of maximizing return and adhering to safety constraints, leading to enhanced sample efficiency. As a result, the training time is reduced, and the generation of unsafe trajectories is minimized.

% In this study, we motivate the introduction of the safety critic and provide a theoretical analysis inspired by the trust region method \cite{trpo}. The Lagrange multiplier method is a special case of our algorithm when specific hyperparameters are chosen. The effectiveness of our algorithm is empirically demonstrated by benchmarking it against five strong and popular safe RL baselines. We use the ball agent from safety bullet gym \cite{BulletSafetyGym} on four tasks: circle, reach, gather, run. 

In this study, we present the rationale for introducing the safety critic and offer a theoretical analysis inspired by the trust region method \cite{trpo}. When particular hyperparameters are selected, the Lagrange multiplier method becomes a special case of our algorithm. The efficacy of our proposed algorithm is empirically validated by benchmarking it against five strong and widely-used safe RL baselines. We employ the ball agent from Safety Bullet Gym \cite{BulletSafetyGym} to evaluate our algorithm on four tasks: circle, reach, gather, and run.

\vspace{-10pt}
\section{Related work}
% By defining a discounted cumulative cost, safe RL aims to ensure safety constraint satisfaction while maximizing return. Most methods utilize Lagrangian relaxation to transform the constrained problem it its unconstrained counterpart. For instance, Ray et al. \cite{Ray} utilizes the update rules from  Trust Region Policy Optimization (TRPO) \cite{trpo} and Proximal Policy Optimization PPO \cite{ppo} to obtain two Lagrangian-based safe RL algorithms named TRPO-Lagrangian and PPO-Lagrangian. Furthermore, they provide a safe RL environment implementation named safety gym, which was used to demonstrate the validity of the proposed algorithms.
% Despite the widespread use of Lagrangian relaxation in solving the safe RL problem, some approaches choose not to utilize this method. For example, Joshua Achiam et al. \cite{cpo} introduce the CPO algorithm, which approximates the constrained optimization problem using a quadratic constrained optimization to handle the constraints. Nonetheless, the computational cost of CPO exceeds that of PPO-Lagrangian because it involves the computation of the Fisher information matrix and utilizes the second Taylor expansion to optimize objectives. Furthermore, the approximation and sampling errors associated with CPO may adversely affect the overall performance, and the convergence analysis might be challenging. In addition, implementing an additional recovery policy may require a larger number of samples.

Safe RL aims to ensure safety constraint satisfaction while maximizing return by defining a discounted cumulative cost. The majority of methods employ Lagrangian relaxation to transform the constrained problem into its unconstrained counterpart. For example, Ray et al. \cite{Ray} leverage the update rules from Trust Region Policy Optimization (TRPO) \cite{trpo} and Proximal Policy Optimization (PPO) \cite{ppo} to derive two Lagrangian-based safe RL algorithms, namely TRPO-Lagrangian and PPO-Lagrangian. They also introduce a safe RL environment implementation called Safety Gym, which is utilized to demonstrate the validity of the proposed algorithms. Although Lagrangian relaxation is commonly used to address the safe RL problem, some approaches opt not to rely on this method. For instance, Joshua Achiam et al. \cite{cpo} propose the Constrained Policy Optimization (CPO) algorithm, which approximates the constrained optimization problem using quadratic constrained optimization to manage the constraints. However, the computational cost of CPO surpasses that of PPO-Lagrangian, as it involves calculating the Fisher information matrix and employing the second Taylor expansion to optimize objectives. Moreover, the approximation and sampling errors associated with CPO can negatively impact the overall performance, and convergence analysis may prove challenging. Additionally, implementing a separate recovery policy might necessitate a larger sample size.

% Inspired by CPO, Projection-based Constrained Policy Optimisation (PCPO) \cite{PCPO} is a 2-stage algorithm. It employs TRPO \cite{trpo} to maximize the reward and then projects the policy to a feasible region to satisfy the safety constraints. Nonetheless, second-order proximal optimization is used in both steps, which increases the computation cost of this algorithm. 
% Different from CPO~\cite{cpo} and PCPO~\cite{PCPO}, Pham et al. introduce a technique called OptLayer \cite{OptLayer}, in which they leverage stochastic control policies to maximize rewards, while integrating a neural network layer to ensure safety during deployment. The practical applications of OptLayer show promising results in enhancing safety. Similar to OptLayer \cite{OptLayer}, a state-augmented safe RL approach, A-CRL \cite{A-CRL}, presents a solution for a CMDP problem in which the optimal policy cannot be obtained solely through regular rewards. The proposed method aims to solve the monitor problem in CMDP, and dual gradient descent is utilized to identify feasible trajectories and ensure safety. However, convergence analysis is not yet provided for A-CRL and OptLayer. 

Drawing inspiration from CPO, Projection-based Constrained Policy Optimization (PCPO) \cite{PCPO} is a two-stage algorithm. It employs TRPO \cite{trpo} to maximize the reward and subsequently projects the policy onto a feasible region to satisfy safety constraints. However, second-order proximal optimization is leveraged in both steps, increasing the computational cost of this algorithm. In contrast to CPO \cite{cpo} and PCPO \cite{PCPO}, Pham et al. introduce a technique called OptLayer \cite{OptLayer}, which leverages stochastic control policies to maximize rewards while incorporating a neural network layer to ensure safety during deployment. The practical applications of OptLayer demonstrate promising results in improving safety. Analogous to OptLayer \cite{OptLayer}, a state-augmented safe RL approach, A-CRL \cite{A-CRL}, presents a solution for CMDP problems where the optimal policy cannot be derived solely through regular rewards. The proposed method aims to address the monitoring problem in CMDP, employing dual gradient descent to identify feasible trajectories and ensure safety. However, convergence analysis for A-CRL and OptLayer has not yet been provided.

Apart from constrained policy optimization methods,  formal methods for safe RL \cite{fulton2018safe} are developed. For example, Hasanbeig et al. \cite{LTL} proposed a safe RL approach that employs reward shaping and linear temporal logic (LTL). This method guarantees safety during exploration by synthesizing policies that satisfy LTL constraints. The LTL formula serves as a constraint during exploration, enabling the search for safe policies. Although it exhibits remarkable safety performance, determining the logical constraints is essential to balance the trade-off between safety performance and reward values.

In addition to CMDP optimization and formal methods for safe RL, control theory is also applied to address safe RL problems. For instance, the Lyapunov function \cite{perkins2002lyapunov} is a popular approach for solving safe RL. It constrains the agent's actions by implementing the control law of Lyapunov functions, removing unsafe actions from the action set. Experiments using this method have demonstrated that it can effectively generate safe actions for control problems \cite{lyapunov2002}. Moreover, Yinlam Chow et al. \cite{lyapunovcontinous} leverage the Lyapunov approach to propose two classes of policy optimization algorithms for continuous tasks, specifically, $\theta$-projection and $\alpha$-projection. However, the Lyapunov approach necessitates an initial feasible policy \cite{lyapunov2018}. While this initial feasible policy can converge to the optimal policy under strict restrictions, creating such a policy can be challenging. Consequently, the Lyapunov approach must be used in conjunction with another optimizer. This requirement limits the generality of the Lyapunov approach as a comprehensive solution to safe RL problems. We provide the proofs of the Lyapunov approach's performance, please refer to Appendix \ref{appendix:Lyapunov-Approach-Analysis}.

% !TeX root = ../main.tex
% Add the above to each chapter to make compiling the PDF easier in some editors.

\section{Preliminaries}\label{chapter:Preliminaries}
To model safe reinforcement learning problems, we use the well-studied framework of constrained Markov decision process (CMDP). A Markov decision process (MDP) is a tuple $(\mathcal{S}, \mathcal{A}, r, P, s_0)$ where $\mathcal{S}$ is the set of states, $\mathcal{A}$ is the action space, $r: \mathcal{S} \times \mathcal{A} \rightarrow \mathbb{R}$ is the immediate reward function, $P: \mathcal{S} \times \mathcal{A} \times \mathcal{S}\rightarrow [0, 1]$ is the environment transition probability distribution and $s_0$ is the initial state. A CMDP extends on top of an MDP and is defined as $(\mathcal{S}, \mathcal{A}, r, c, P, s_0, c_0)$, where $c: \mathcal{S} \rightarrow \mathbb{R_+}$ is the immediate cost function and $c_0$ is the maximum allowed cumulative cost. \\
Let $\Delta = \{ \pi: \mathcal{S}, \mathcal{A} \rightarrow [0, 1], \  \forall{s}\in\mathcal{S} \quad \sum_{a\in\mathcal{A}} \pi(a, s) = 1\}$ be the set of all policies.
Given a policy $\pi$, we define the expected cumulative cost, $\mathcal{C}_\pi(s_0) = \EX_\pi\left[ \sum_{t=0} \gamma^t c(S_t), \  S_0=s_0\right]$, 
% \begin{align}
%     \mathcal{C}_\pi(s_0) = \EX_\pi\left[ \sum_{t=0} \gamma^t c(S_t), \  S_0=s_0\right] \nonumber
% \end{align}
The safety constraint is defined as $\mathcal{C}(s_0) \leq c_0$. The CMDP goal is to solve the following constrained optimization problem, $\pi^\ast \in \max_{\pi\in\Delta}\{ V_\pi(s_0), \ \mathcal{C}_\pi(s_0) \leq c_0 \}$.
% \begin{align}
%     \pi^\ast \in \max_{\pi\in\Delta}\{ V_\pi(s_0), \ \mathcal{C}_\pi(s_0) \leq c_0 \} \nonumber
% \end{align}

\section{Method}
\subsection{Safety critic}
% !TeX root = ../main.tex
% Add the above to each chapter to make compiling the PDF easier in some editors.
 \label{chapter:Safety_metric}
This section introduces the safety critic $V^c(s)$, which represents the probability of generating an unsafe trajectory from a given state $s$. We also define $Q^c(s, a)$ as the probability of generating an unsafe trajectory from state $s$ after taking action $a$. These functions enable the efficient search for an optimal safe policy. The rationale is as follows: The agent is in a state $s$, selecting action $a_1$ leads to a considerably high reward, but violates the safety constraint. Alternatively, selecting action $a_1^\prime$ does not violate the constraint but offers significantly less reward. The agent must balance two objectives: maximizing return and minimizing cost.
We argue that the reward obtained by choosing the unsafe action $a_1$ should be nullified. It is counterproductive to take into consideration rewards obtained by acting in an unsafe manner. By changing our reward function to $r^\prime(s, a) = r(s, a) \ Q^c(s, a)$, we decrease the effect of unsafe actions. If $Q^c(s, a_1) = 0$, indicating that $a_1$ always violates safety constraints, the agent is not incentivized to select it. In contrast, if $Q(s, a_1^\prime) = 1$, the reward obtained from choosing $a_1^\prime$ is unchanged, e.g. $r^\prime(s, a_1^\prime) = Q^c(s, a_1^\prime) r(s, a_1^\prime) = r(s, a_1^\prime)$.
\subsection{Augmented States with Cumulative Cost}
\label{section:augment_state}
\begin{figure}[!h]
     \centering
     \includegraphics[width=0.86\linewidth]{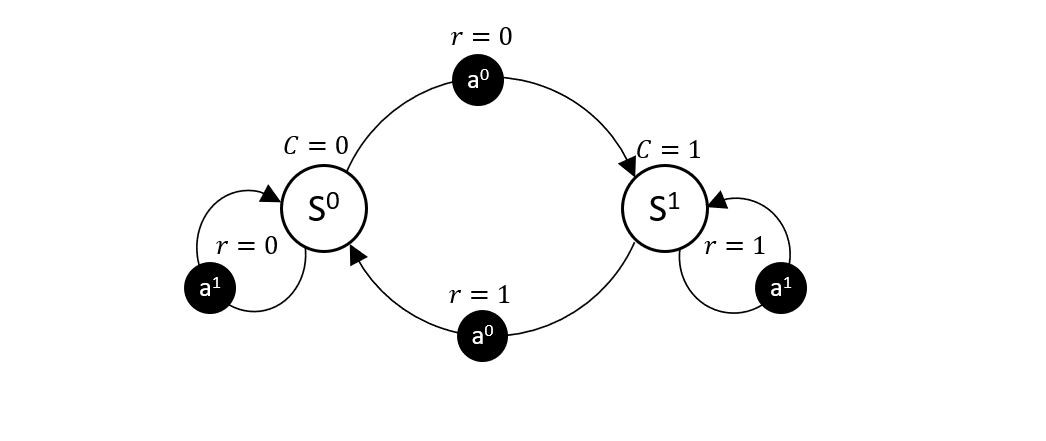}
     \caption{CMDP where the optimal policy cannot be found if the state representation does not contain the cumulative cost; the agent has to guess when the constraint violation occurs. When the cumulative cost is included in the state representation, the agent can make more informed decisions and reach a deterministic policy, e.g. $\pi(a \kgiven s) = 0 \text{ or } 1$.}
     \label{CMD:augment}
\end{figure}

Incorporating safety into reinforcement learning necessitates that the agent behaves differently based on the current cumulative cost. Specifically, the agent can take actions that increase the cumulative cost if the maximum cumulative cost $c_0$ has not been exceeded.
To provide a concrete example of this principle, we define the following CMDP, as shown in Figure \ref{CMD:augment}, with two states $s^0$, $s^1$ and two actions $a^0$, $a^1$. The episode length is 10 and $c_0 = 5$.
Let us consider the case where $s_t = \mathbbm{1}[s_t = s^1]$. To behave optimally, the agent should select action $a^0$ and remain in state $s^1$ for five time steps by taking action $a^1$. At $t=5$, the maximum cumulative cost constraint is reached. To avoid violating the safety constraint, the agent must select action $a^0$, transitioning to state $s^0$. However, this decision cannot be made because it lacks information regarding the current cumulative cost.\\
To address this issue, we introduce a new variable $q_t$ representing the cumulative cost at time $t$. By including $q_t$ in the state representation, we can solve the problem discussed above. We define the new state as $s_t^\prime = (\mathbbm{1}[s_t = S^1], q_t)$. However, this increases the number of possible states from 2 to 20.
When dealing with continuous tasks, the cumulative cost might be unbounded. Hence, we introduce $q_t^\text{clip} = \text{clip}(\frac{q_t}{c_0}, 0, 1+\Delta)$ where $\Delta \approx 0$. $\frac{q_t}{c_0}$ is clipped because the agent's behavior should remain the same after reaching the maximum allowed cumulative cost. In the upcoming sections, we assume all states are augmented using $q^{\text{clip}}_t$.
\subsubsection{Safety critic definition}
For any state $s \in \mathcal{S}$ and trajectory, we define the function $f: \mathcal{S} \rightarrow [0, 1]$ as follows:
\begin{equation*}
    f(s) = \begin{cases}
    1  \qquad \text{if $s$ is safe, $q^{\text{clip}} \leq 1$},\\
    0  \qquad \text{if $s$ is unsafe, $q^{\text{clip}} > 1$}.\\
    \end{cases}
\end{equation*}
For a trajectory $\tau = (s_0, a_0, s_1, a_1 ... s_{T-1})$, we define $f(\tau) = \prod_{t=0}^{T-1} f(s_t)$. We call the trajectory safe if $f(\tau) = 1$.
{
\definition{}{}
For any $s\in\mathcal{S}$, action $a$ and policy $\pi$
\begin{equation*}
    V^c_\pi(s) = \EX_\pi\left[\prod_{t=0}^{T-1} f(S_t) \kgiven S_0=s \right], \qquad  Q^c_\pi(s, a) = \EX_\pi\left[\prod_{t=0}^{T-1} f(S_t) \kgiven S_0=s, A_0=a \right].
\end{equation*}
}
\\
$V_\pi^c(s)$ is the probability of generating a safe trajectory starting from state s:\\  $V_\pi^c(s) = \sum_{\tau} P_\pi[\tau \kgiven S_0=s] \ f(\tau)$.
\begin{theorem}
 \label{theorem:v_c_diff_Q}
\begin{equation*}
     \nabla V^c_\pi(s) = \EX_\pi\left[\sum_{t=0} \nabla\log(\pi(A_t \kgiven S_t)) Q_\pi^c(S_t, A_t) \kgiven S_0=s \right].
\end{equation*}
\end{theorem}
Proof: See Appendix \hyperref[proof:vc-diff-Q]{A}.

For an arbitrary state $s\in\mathcal{S}$ and action $a\in \mathcal{A}$ we defined the safety advantage $A^c_\pi(a, s) = Q^c_\pi(s,a) - V^c_\pi(s)$.
\begin{corollary}
\label{corollary:v_c_diff_A}
\begin{equation*}
     \nabla V^c_\pi(s) = \EX_\pi\left[\sum_{t=0} \nabla\log(\pi(A_t \kgiven S_t)) A_\pi^c(S_t, A_t) \kgiven S_0=s \right].
\end{equation*}
\end{corollary}
Proof: See Appendix \hyperref[proof:v-c-diff-A]{A}.\\
Approaches for estimating $Q^c$ and $V^c$ are discussed in detail in Appendix~\ref{appendix-section:V-C-ESTIMATE}.
\subsubsection{General trust region results} \label{chapter:trust_region}
{
\begin{theorem}
\label{thorem:trust-region-1}
For arbitrary policies $\pi$ and $\pi^\prime$ and state $s$, and $\gamma \in [0, 1]$  the following equality holds:
\begin{align}
    \label{eq-method-main:V_V_C}
    &V_{\pi^\prime}(s) = V_\pi(s) + \EX_{\pi^\prime}\left[\sum_{t=0} \gamma^t (1-\frac{\pi(A_t \kgiven S_t)}{\pi^\prime(A_t \kgiven S_t)}) Q_\pi(S_t, A_t) \kgiven S_0=s\right], \\
    &V_{\pi^\prime}^c(s) = V_\pi^c(s) + \EX_{\pi^\prime}\left[\sum_{t=0} (1-\frac{\pi(A_t \kgiven S_t)}{\pi^\prime(A_t \kgiven S_t)}) Q_\pi^c(S_t, A_t) \kgiven S_0=s\right].
\end{align}
\end{theorem}
}
Proof: See Appendix \hyperref[proof:trust-region-1]{ A}.

Maximizing $V_{\piprime}(s)$ is therefore equivalent to maximizing 

\begin{center}
    \scalebox{0.99}{$\EX_{\pi^\prime}\left[\sum_{t=0} (1-\frac{\pi(A_t, S_t)}{\pi^\prime(A_t, S_t)}) Q_\pi^c(S_t, A_t)\kgiven S_0=s\right]$}.
\end{center} 

However, evaluating this term requires sampling trajectories using $\piprime$. This necessitates discarding the data sampled using $\pi$ and generating a new one using $\pi^\prime$ after one epoch or iteration. Following the approach in \cite{trpo}, we instead want to find a term that maximises $V_{\piprime}(s)$ while using samples generated by $\pi$.
In the following, we prove properties with respect to $V$. The results can be generalized to $V^c$ because only common algebraic properties were used.

\begin{corollary}
\label{corollary:trust-region-2}
\begin{equation}
\label{Q_TO_A}
    V_{\pi^\prime}(s) = V_\pi(s) + \EX_{\pi^\prime}\left[\sum_{t=0} \gamma^t (1-\frac{\pi(A_t\kgiven S_t)}{\pi^\prime(A_t\kgiven S_t)}) A_\pi(S_t, A_t) \kgiven S_0=s\right].
\end{equation}
\end{corollary}
Proof: See Appendix \hyperref[proof:trust-region-2]{A}.
\begin{theorem}
\label{thorem:trust-region-estimator-1}
Let $\alpha = \max_s D_{TV}(\pi^\prime(a, .), \pi(a, .))$, 
$\epsilon = \max_{s, a} \rvert (\frac{\pi^\prime(a \kgiven s)}{\pi(s\kgiven a)} - 1) A_\pi(s, a) \lvert$ and $\gamma\in [0, 1)$,
\begin{equation}
    V_{\piprime}(s) \ge V_\pi(s) + \EX_{\pi}\left[\sum_{t=0} \gamma^t (\frac{\pi^\prime(A_t \kgiven S_t)}{\pi(A_t \kgiven S_t)} -1) \ A_\pi(S_t, A_t) \kgiven S_0=s\right] - \frac{2 \alpha \gamma \epsilon}{(1-\gamma)^2}.
\end{equation}
\end{theorem}
Proof: see Appendix \hyperref[proof:trust-region-estimator-1]{A}.

The above Theorem \ref{thorem:trust-region-estimator-1} suggests that in addition to constraining \scalebox{0.75}{$\max_s D_{TV}(\pi^\prime(a, .), \pi(a, .))$}, it is also important to constrain the relative distance, $\max_{s, a} |\frac{\pi^\prime(a, s) - \pi(a, s)}{\pi^\prime(a, s)}|$. The proximal policy optimization (PPO) \cite{ppo} addresses this point by clipping the ratio $\frac{\pi^\prime(a, s)}{\pi^\prime(a, s)}$ to be close to $1$, which constrains $\frac{\pi^\prime(a \kgiven s)}{\pi^\prime(a \kgiven s)} - 1$.
Furthermore, to maximize $V_{\piprime}(s_0)$, we maximize $\EX_{\pi}\left[\sum_{t=0} \gamma^t (\frac{\pi^\prime(A_t, S_t)}{\pi(A_t, S_t)} -1) \ A_\pi(S_t, A_t) \kgiven S_0=s\right]$ while ensuring that $\frac{2 \alpha \gamma \epsilon}{(1-\gamma)^2}$ is close to zero. 
We use the update rule from PPO \cite{ppo} because the gradient is the same, and the constraints are similar to \cite{trpo}:
 \begin{align*}
    \nabla \EX_{\pi}\left[\sum_{t=0} \gamma^t (\frac{\pi^\prime(A_t, S_t)}{\pi(A_t, S_t)} -1) \ A_\pi(S_t, A_t) \kgiven S_0=s\right]
    \\
    = \nabla \EX_{\pi}\left[\sum_{t=0} \gamma^t \frac{\pi^\prime(A_t, S_t)}{\pi(A_t, S_t)} \ A_\pi(S_t, A_t) \kgiven S_0=s\right].
\end{align*}

Instead of evaluating $\EX_{\pi^\prime}\left[\sum_{t=0} \gamma^t (1-\frac{\pi(A_t \kgiven S_t)}{\pi^\prime(A_t \kgiven S_t)}) A_\pi(S_t, A_t)\right]$, which requires sampling trajectories using $\pi^\prime$, we can instead evaluate 

\begin{center}
    $\EX_{\pi}\left[\sum_{t=0} \gamma^t(1 - \frac{\pi(A_t \kgiven S_t)}{\pi^\prime(A_t \kgiven S_t)}) A_\pi(S_t, A_t)\right]$,
\end{center}
and examine the difference between the two.

\begin{theorem}    
Let $\alpha = \max_s D_{TV}(\pi^\prime(a, .), \pi(a, .))$,  $\epsilon^\prime = \max_{s, a} \rvert (1 -\frac{\pi(s \kgiven a)}{\pi^\prime(s \kgiven a)}) A_\pi(s, a) \lvert$ and $\gamma\in [0, 1)$,
\label{theorem:trust-region-estimator-2}
\begin{equation}
 V_{\piprime}(s) \ge V_\pi(s) + \EX_{\pi}\left[\sum_{t=0} \gamma^t(1 - \frac{\pi(A_t \kgiven S_t)}{\pi^\prime(A_t \kgiven S_t)}) A_\pi(S_t, A_t) \kgiven S_0= s  \right] - \frac{2 \alpha \gamma \epsilon^\prime}{(1-\gamma)^2}.
\end{equation}
\end{theorem}
Proof: see Appendix \hyperref[proof:trust-region-estimator-2]{A}.

The \textbf{differences between the equations of Theorem \ref{thorem:trust-region-estimator-1} and Theorem \ref{theorem:trust-region-estimator-2}} are provided in the Appendix~\ref{appendix-sec:analyzing-the-difference-setimators}.

\subsection{Safe Policy Iteration}\label{chapter:SafePolicyIteration}
In this section, we use $Q^c(s, a)$ to cancel unsafe rewards. We gradually modify the value function $V_\pi(s_0) = \left[\sum_{t=0} r(S_t, A_t) \kgiven S_0 =s_0\right]$ to cancel unsafe returns and give motivating examples. We assume that the reward function is positive:
$\forall{s\in\mathcal{S}\ \forall{a\in\mathcal{A}}}, r(s, a) \geq 0$.
\subsubsection{Canceling Unsafe Reward}
\label{subsec:Canceling-Unsafe-Reward}
Let $\tau=(s_0, a_0, s1, a_1.., a_{T-1}, s_{T-1})$ be a trajectory and $R(\tau) = \sum_t \gamma^t r(s_t, a_t)$.  We define $V^r: \mathcal{S}, \mathcal{A}  \rightarrow \mathbb{R}$ as follows:
\begin{equation*}
    V^r_\pi(s) = \sum_{\tau} P_\pi[\tau \kgiven S_0=s] \ R(\tau) \ f(\tau).
\end{equation*}
If a trajectory $\tau$ is unsafe, then $R(\tau) \ f(\tau) = 0$. Therefore, $\nabla P_\pi[\tau] \ R(\tau) \ f(\tau) = 0$. In other words, we don't increase the probability of generating $\tau$ using $\pi$ when we follow $\nabla V^r_\pi$. However, if $f(\tau) = 1$, we increase the probability of generating $\tau$ proportionally with $R(\tau)$.

\begin{equation}
\label{V_R}
    V^r_\pi(s) = \EX_\pi\left[\sum_{t=0} \gamma^t r(S_t, A_t) \ Q^c_\pi(S_t, A_t) \kgiven S_0=s \right].
\end{equation}
Proof: see Appendix \hyperref[proof:V_R]{A}.

We follow the standard definition for $Q^r_\pi: \mathcal{S}, \mathcal{A} \rightarrow \mathbb{R}$ and $A^r_\pi: \mathcal{S}, \mathcal{A} \rightarrow \mathbb{R}$:
\begin{align*}
    Q^r_\pi(s, a) &= \EX_\pi\left[\sum_{t=0} \gamma^t r(S_t, A_t) Q^c_\pi(S_t, A_t) \kgiven S_0=s, A_0=a\right] ,\\
    A^r_\pi(s, a) &= Q^r_\pi(s, a) - V^r_\pi(s) .
\end{align*}
$V^r_\pi$ can be viewed as a value function with reward $r^\prime(s, a) = r(s, a) \ Q^c_\pi(s, a)$.

\begin{figure}[H]
    \centering
    \includegraphics[width=0.85\textwidth]{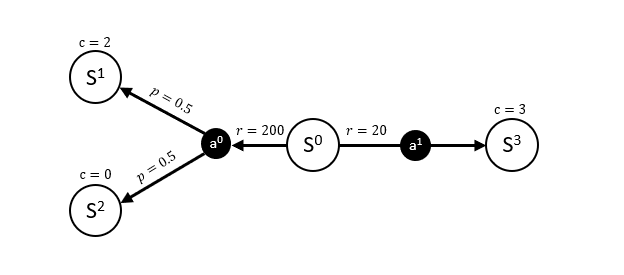}
    \caption{CMDP where optimizing w.r.t $V^r$ leads to an unsafe policy.}
    \label{CMDP:v_r}
\end{figure}

We note that $ V_\pi(s_0) \ge V_\pi^r(s_0)$ and the equality holds if $V_\pi^c(s_0) = 1$.
Let's consider the 3 state CMDP, as shown in Figure \ref{CMDP:v_r}, where the maximum cumulative cost $c_0=1$.
If the agent chooses action $a^0$ from the initial state $s^0$, it transitions to $s^1$ or $s^2$ with the following transition probability: $p(s^1\kgiven s^0, a^0) = 0.5$ and $p(s^2\kgiven s^0, a^0) = 0.5$. The agent chooses both $a^0$ or $a^1$ with probability $0.5$\\
\begin{alignat*}{2}
    &Q_\pi^c(s^0, a^0) = 0.5; \qquad &&A_\pi^r(s^0, a^0)=40 ;\\
    &Q_\pi^c(s^0, a^1) = 1 ;\qquad  &&A_\pi^r(s^0, a^1)=-40.
\end{alignat*}
Because $Q_\pi(s^0, a^0) \gg Q_\pi(s^0, a^1)$, updating the policy can increase the probability of choosing $a^0$. However, the reward obtained from choosing $a^0$ should be canceled because $Q_\pi^c(s^0, a^0) = 0.5 < 1$; picking action $a^0$ violates the safety constraint with probability $0.5$. Therefore, it is reasonable to cancel the reward of actions that violate the safety constraints. We can formalize this property as follows: $V_\pi^{r, k} = \EX_\pi\left[\sum_{t=0} \gamma^t r(S_t, A_t) \ (Q^c_\pi(S_t, A_t))^k \kgiven S_0=s \right]$.
Taking the limit of $k$ as it approaches infinity:
\begin{align*}
    \lim_{k \rightarrow \infty}  V_\pi^{r, k} &= \EX_\pi\left[\sum_{t=0} \gamma^t r(S_t, A_t) \ \vert\vert Q^c_\pi(S_t, A_t))\vert\vert_{\infty} \kgiven S_0=s \right]\\
    &= \EX_\pi\left[\sum_{t=0} \gamma^t r(S_t, A_t) \  \mathbf{I}[Q^c_\pi(S_t, A_t) = 1] \kgiven S_0=s \right].
\end{align*}
Computing different values of $A_\pi^{r, k}(s,a)$ of the CMDP \ref{CMDP:v_r}:
\begin{alignat*}{2}
    A_\pi^{r, 0}(s^0, a^0) &= 90; \qquad \qquad &&A_\pi^{r, 0}(s^0, a^1) = -90;\\
    A_\pi^{r, 1}(s^0, a^0) &= 40; \qquad \qquad &&A_\pi^{r, 1}(s^0, a^1) = -40;\\
    A_\pi^{r, 4}(s^0, a^0) &= -3.75; \qquad \qquad &&A_\pi^{r, 0}(s^0, a^1) = 3.75;\\
    A_\pi^{r, 8}(s^0, a^0) &= -9.6; \qquad \qquad &&A_\pi^{r, 8}(s^0, a^1) = 9.6;\\
    A_\pi^{r, \infty}(s^0, a^0) &= -10; \qquad \qquad &&A_\pi^{r, \infty}(s^0, a^1) = 10.\\
\end{alignat*}
As demonstrated by this example, it is not always necessary to choose $k \rightarrow \infty$. Instead, choosing $k=4$ solves this toy example.\\
Because we restrict the reward function $r$ to be positive and $Q^c_\pi(s, a) \in [0, 1]$, the following property holds:
$V_\pi^{r, 0} = V_\pi \geq V_\pi^{r, 1} \geq V_\pi^{r, 2} \dots \geq V_\pi^{r, \infty}$.
\begin{restatable}{theorem}{} \textbf{Stability Analysis}.
Let $r \geq 0$. The following holds:
\begin{equation}
    \pi^\ast = \argmax_{\pi} V_\pi^{r, \infty}(s_0) \Rightarrow V^c_{\pi^\ast}(s_0) = 1.
\end{equation}
\end{restatable}
Because the only way to obtain reward is to behave in a safe manner, maximizing $V_\pi^{r, \infty}(s_0)$ leads to a safe policy.
We use 
\begin{center}
    $\bar{V}_{\pi^\prime}^{r,k} = \EX_{\pi^\prime}\left[\sum_{t=0} \gamma^t r(S_t, A_t) Q_\pi^c(S_t, A_t)^k\kgiven S_0=s_0\right]$ 
\end{center}
as a first order approximation of $V_{\pi^\prime}^{r,k}$ when $\pi \approx \pi^\prime$.
\begin{theorem}
\label{V-rc-bar}
\begin{equation}
    \bar{V}_{\pi^\prime}^{r,k}(s_0) = V_{\pi}^{r,k}(s_0) + \EX_{\pi^\prime}\left[\sum_{t=0} \gamma^t \ A_\pi^{r, k}(S_t, A_t) \kgiven S_0=s_0\right].
\end{equation}
\end{theorem}
Proof: see Appendix \hyperref[proof:V-rc-bar]{A}. 
\begin{theorem}
\textbf{Convergence Analysis}.
Let $\alpha = \max_s D_{TV}(\pi^\prime(a\kgiven .), \pi(a\kgiven .))$, $\gamma\in [0, 1)$,\\
$\epsilon = \max_{s, a} \rvert (\frac{\pi^\prime(s \kgiven a)}{\pi(s \kgiven a)} - 1) A_\pi^{r, k}(s, a) \lvert$, $\epsilon^\prime = \max_{s, a} \rvert (1 -\frac{\pi(s \kgiven a)}{\pi^\prime(s \kgiven a)}) A_\pi^{r, k}(s, a) \lvert$,
\begin{align}
    \bar{V}_{\piprime}^{r,k}(s) \ge V_\pi^{r,k}(s) + \EX_{\pi}\left[\sum_{t=0} \gamma^t (\frac{\pi^\prime(A_t \kgiven S_t)}{\pi(A_t\kgiven S_t)} -1) \ A_\pi^{r,k}(S_t, A_t) \kgiven S_0=s\right] - \frac{2 \alpha \gamma \epsilon}{(1-\gamma)^2},\\
     \bar{V}_{\piprime}^{r,k}(s) \ge V_\pi^{r,k}(s) + \EX_{\pi}\left[\sum_{t=0} \gamma^t(1 - \frac{\pi(A_t\kgiven S_t)}{\pi^\prime(A_t\kgiven S_t)}) A_\pi^{r,k}(S_t, A_t) \kgiven S_0= s  \right] - \frac{2 \alpha \gamma \epsilon^\prime}{(1-\gamma)^2}.
\end{align}
\end{theorem}
Proof omitted due to similarity with Theorem \ref{thorem:trust-region-estimator-1} and Theorem \ref{theorem:trust-region-estimator-2}.\\
We also note that choosing an appropriate value of $k$ can solve problems where $V_\pi^\ast(s_0) < 1$. To illustrate this, consider the following CMDP:
\begin{figure}[H]
     \centering
     \includegraphics[width=0.85\textwidth]{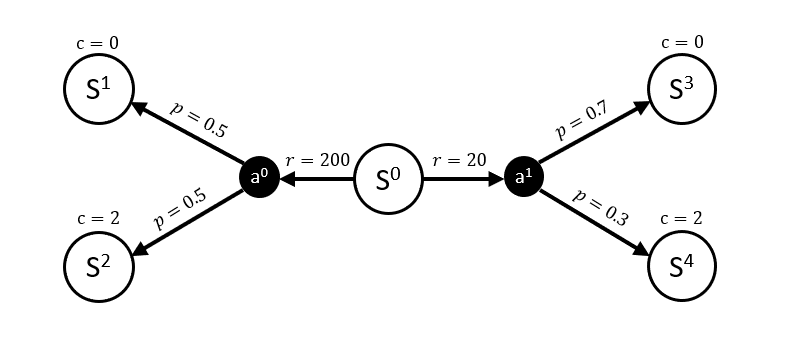}
     \caption{CMDP where all policies are not totally safe, e.g $V^c_\pi(s^0) < 1$.}
     \label{CMDP:not-totally-safe}
\end{figure}
The maximum achievable safety is $V_{\pi^\ast}^c(s^0) = 0.7$. We compute $A_\pi^{r, k}(s, a)$ with different $k$ values:
\begin{alignat*}{2}
    A_\pi^{r, 0}(s^0, a^0) &= 90; \qquad \qquad &&A_\pi^{r, 0}(s^0, a^1) = -90;\\
    A_\pi^{r, 4}(s^0, a^0) &= 3.84; \qquad \qquad &&A_\pi^{r, 4}(s^0, a^1) = -3.84;\\
    A_\pi^{r, 8}(s^0, a^0) &= -0.18; \qquad \qquad &&A_\pi^{r, 8}(s^0, a^1) = 0.18;\\
    A_\pi^{r, \infty}(s^0, a^0) &= 0; \qquad \qquad &&A_\pi^{r, \infty}(s^0, a^1) = 0.\\
    \end{alignat*}
Because $V_{\pi^\ast}(s^0) < 0$, $A_\pi^{r, \infty}(s, a) = 0$. Choosing $k=8$ achieves the best trade-off between safety and return maximization.

% \textbf{Reducing Cost:}
% \subsubsubsection{Reducing Cost}
% \label{section:reducing-cost}
The safety metric $V^c(s)$ reflects the probability of visiting safe states starting from $s$. Let $a^1$ and $a^2$ be two actions such that $Q^c(s^0, a^1) = Q^c(s^0, a^2) = 0.9$. Even though both actions are equally safe, it is possible that $E_\pi[\sum_t c(S_t) \kgiven S_0=s, A_0=a_1] \gg E[\sum_t c(S_t) \kgiven S_0=s, A_0=a_2]$. Therefore, it is desirable to favour $a^2$ over $a^1$. The CMDP in figure \ref{CMDP:same-safety} illustrates this problem:
\begin{figure}[!ht]
  \centering   \includegraphics[width=0.85\textwidth]{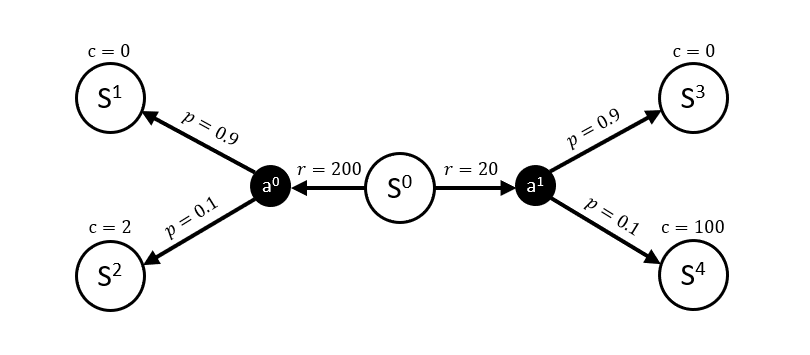}
     \caption{Using the objective function $V^{r,k}$ leads to multiple solutions; Actions $a^0$ and $a^1$ are equally favoured. The optimal solution should also minimize the cumulative cost. This CMDP motivates the introduction of a cost term to $V^{r, k}$.}
     \label{CMDP:same-safety}
\end{figure}
Because $\forall{k \geq 0} \ A^{r,k}_\pi(s^0, a^1) = A^{r,k}_\pi(s^0, a^2)$, our objective function $V^{r, k}_\pi(s^0)$ equally favors both actions. This motivates the introduction of a new term to $V^{r, k}$:
{
\definition{}{} Let $\beta \geq 0$,
\begin{equation}
    V_\pi^{rc, k}(s) = \EX_\pi\left[\sum_{t=0} \gamma^t (r(S_t, A_t) \ Q^c_\pi(S_t, A_t)^k - \beta \ (1-Q^c_\pi(S_t, A_t)^k) \ c(S_t)) \kgiven S_0=s \right].
\end{equation}
}
If $V^c_\pi(s_0) = 1$, then $V_\pi(s_0) = V^{r, k}(s_0) = V^{rc, k}(s_0)$. Therefore, the maximum achievable safe return does not change.
Moreover, 
\begin{center}
    $\EX_\pi\left[\sum_{t=0} \gamma^t (1-Q^c_\pi(S_t, A_t)^k) \ c(S_t) \kgiven S_0=s, A_t=a \right]$ 
\end{center}
allows us to distinguish between unsafe actions based on how much cumulative cost they incur. This facilitates avoiding extremely hazardous behavior.
Using $V^{rc, k}$ is also useful when the randomly initialized policy is entirely unsafe, e.g. $Q^c_\pi(s, a) = 0$ for all state-action pairs. In such scenario, using $V^{rc, k}$ is equivalent to minimizing the cumulative cost:
\begin{equation*}
   \forall{a\in\mathcal{A}} \ \forall{s\in\mathcal{S}}  \quad Q^c_\pi(s, a) = 0 \Rightarrow V^{rc, k}_{\pi}(s_0) = -\EX_\pi\left[\sum_{t=0} \gamma^t c(S_t) \kgiven S_0=s_0 \right].
\end{equation*}
When all generated trajectories are unsafe, minimizing the cumulative cost is necessary to find a safe policy eventually. We define the $Q$ function and advantage with respect to $V^{rc, k}$ as follows:
\begin{align*}
    &Q^{rc, k}_\pi(s, a) =\\& \EX_\pi\left[\sum_{t=0} \gamma^t (r(S_t, A_t) \ Q^c_\pi(S_t, A_t)^k - \beta \ (1-Q^c_\pi(S_t, A_t)^k) \ c(S_t)) \kgiven S_0=s, A_0=a\right], \\
    &A^{rc, k}_\pi(s, a) = Q^{rc, k}_\pi(s, a) - V^{rc, k}_\pi(s) .
\end{align*}
Inspired by GAE~\cite{GAE}, the equation, $\hat{A}^{\text{GAE}(\gamma, \lambda)} = \sum_{l=0} (\gamma \lambda)^l \delta^V_{t+l}$, can be used to estimate $A^{rc, k}_\pi$. For more detail analysis, please see Appendix~\ref{appendix-generalized-advantage-estimation}.
\subsubsection{Practical Implementation}\label{chapter:PraticalImplementation}
% \subsection{Safety critic policy optimization}
After generating trajectories using $\pi$, we can compute $\hat{A}^{r, r}_\pi(s, a)$ and $\hat{A}^c_\pi(s, a)$ for every state-action pair.
We denote a parameterized policy by $\pi_{\theta}$. We refer to the initial policy by $\pi_{\theta_0}$.
To approximate $\pi$, $V^c$, and $V^{rc, k}$, we use three fully-connected MLPs with two hidden layers and tanh nonlinearities. The policy network outputs the mean of a Gaussian distribution with variable standard deviations for continuous tasks, as described in \cite{trpo,dua}.
When the reward is negative, a reward bias is added such that for all state $s$ and action $a$, $r^\prime(s, a) = r(s, a) + b \geq 0$. We start by a randomly initialized policy $\pi_{\theta}$, the practical algorithm and more detail analysis are provided in Algorithm~\ref{appendix-alg:SAFEPPO} and Appendix~\ref{appendix-chapter:PraticalImplementation}.

\section{Experiments}\label{chapter:Experiments}
In this section, we evaluate the performance of our algorithm (SCPO) by comparing it to TRPO-L \cite{Ray}, CPO \cite{cpo}, PDO, and PCPO \cite{PCPO}. We use the ball agent from safety bullet gym \cite{BulletSafetyGym} on four tasks: circle, reach, gather, run. The experiment results demonstrate the effectiveness of our approach. We also highlight the importance of augmenting the state representation using the cumulative cost as described in Section \ref{section:augment_state}, the environmental settings are introduced in Appendix~\ref{appendix-section:environments-settings}, related comparing strong baselines are analyzed in Appendix~\ref{appendix-safe-baselines-introduction} and discrete setting experiments are provided in Appendix~\ref{subsection:CartSafe}.
Figure \ref{figure-experiments:benchmark_bullet} demonstrates that SCPO outperforms all other algorithms. Although the final return is similar for most tasks, SCPO significantly outperforms in the SafetyBallGather task. Furthermore, the cost and cost standard deviation of SCPO are consistently lower than the other algorithms during training.
SCPO is capable of finding a safe policy at a faster rate compared to the other algorithms. The cost promptly drops below the threshold without compromising the improvement of the return. This highlights SCPO's ability to better balance the two primary objectives: satisfying safety constraints and maximizing return.

% Experimental videos are available on the page \url{https://sites.google.com/view/saferl-scpo/home}.

\begin{figure*}
\begin{figure}[H]
    \includegraphics[width=\textwidth]{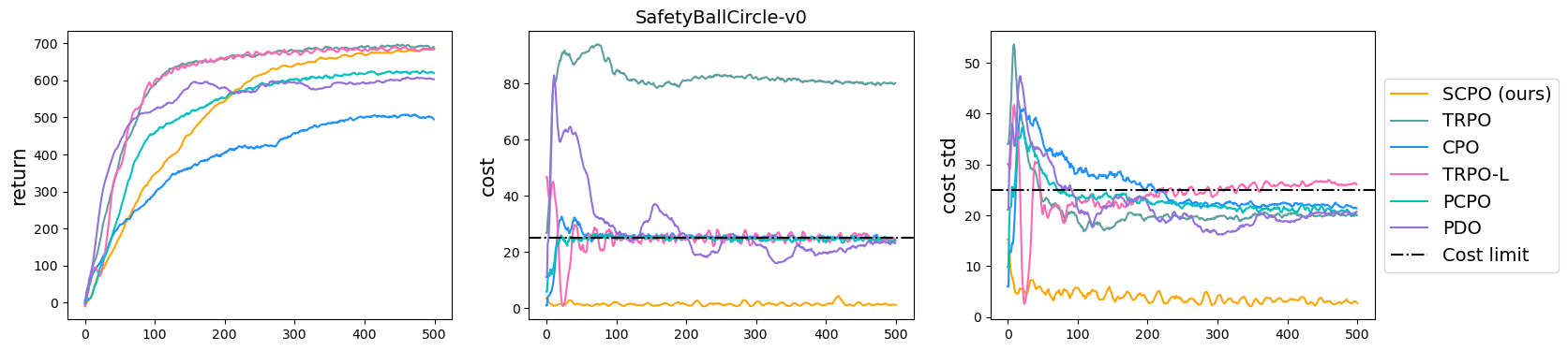}
    \includegraphics[width=\textwidth]{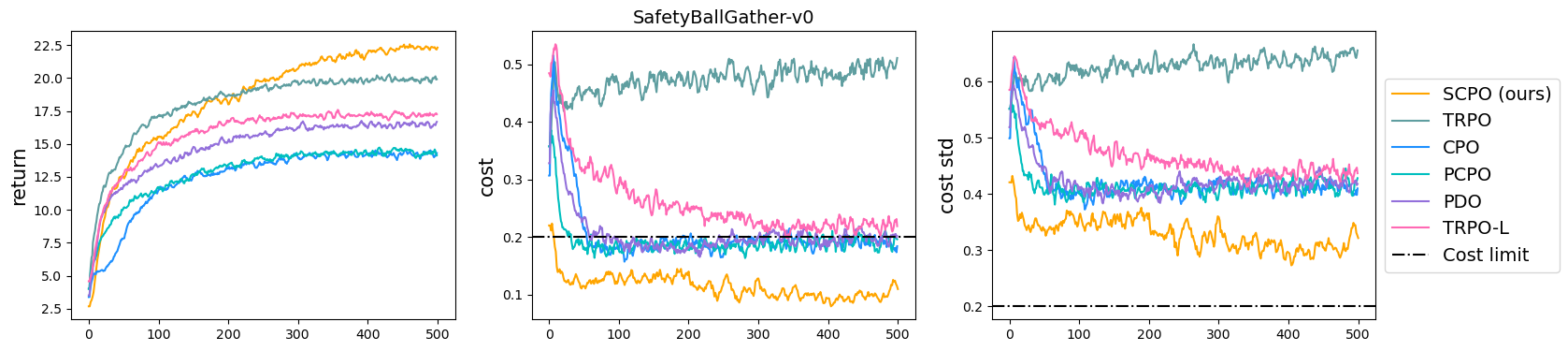}
    \includegraphics[width=\textwidth]{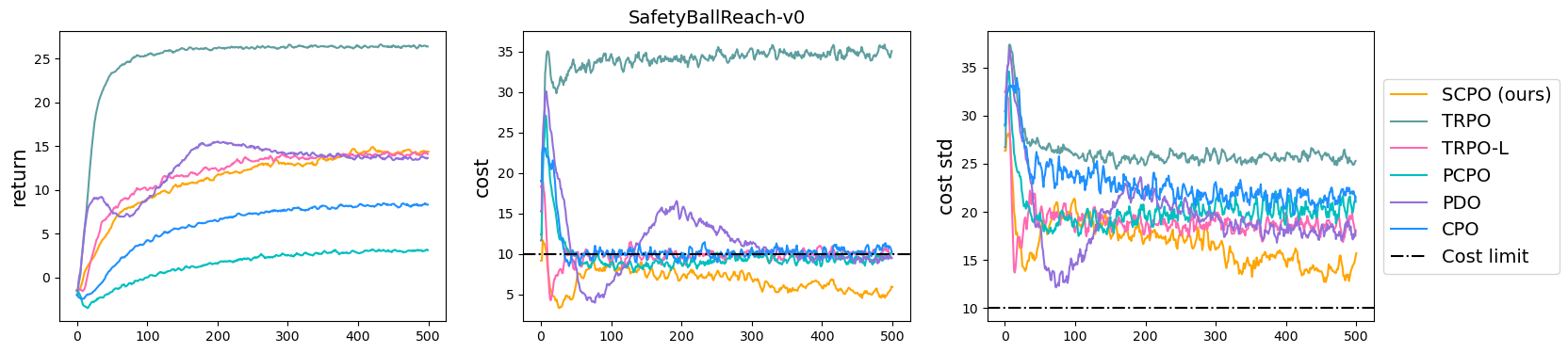}
    \includegraphics[width=\textwidth]{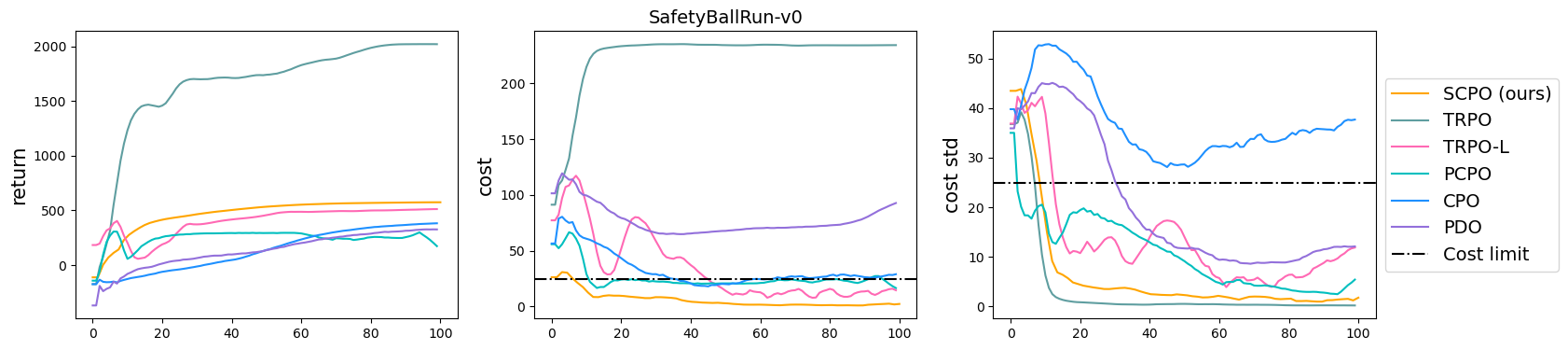}
    \caption{The learning curve of each algorithm average over four different seeds on four tasks: circle, reach, gather, run.}
    \label{figure-experiments:benchmark_bullet}
\end{figure}
\end{figure*}

\section{Conclusion}\label{chapter:conclusion-future}
In this study, we proposed a novel approach to solve safe reinforcement learning problems. Our work introduces the safety critic, which is used to nullify rewards obtained by violating safety constraints. Moreover, safety critic helps manage the trade-off between cost reduction and return maximization. Our approach is straightforward to implement and can be easily integrated with current RL methods. Empirically, we validate our theoretical findings and compare our algorithm (SCPO) to TRPO-L \cite{Ray}, CPO \cite{cpo}, PDO~\cite{ray2019benchmarking}, PCPO \cite{PCPO}, and TRPO~\cite{trpo}. Our approach violates the safety constraint significantly less than the other algorithms throughout training without sacrificing improving the return. It also converges to a safe policy faster than other approaches. Our work is a step forward in deploying RL to real-world problems where safety guarantees are critical. Future research will focus on convergence analysis and evaluating the efficacy of SCPO in complex and challenging environments. Additionally, the neural network architecture of the safety critic can be improved to be statistically more sound.

\bibliographystyle{plain}
\bibliography{main}

\clearpage

\appendix
\begin{center}
\textbf{\Large Appendix}
\end{center}

% !TeX root = ../main.tex
% Add the above to each chapter to make compiling the PDF easier in some editors.

\section{Lyapunov Approach Analysis}\label{appendix:Lyapunov-Approach-Analysis}
\iffalse
\begin{table}[!h]
\begin{center}
\begin{tabular}{  c  c  c  } 
  \hline
  Environment Name & Cost limit & Maximum Episode Length \\
\hline
  SafetyBallCircle-v0   & 25 & 250 \\ 
  SafetyBallRun-v0  & 25 & 250 \\ 
  SafetyBallGather-v0 & 0.2 & 250 \\
  SafetyBallReach-v0 & 10 & 250 \\
  CartSafe-v0 & 1 & $[0, 300]$ \\
  \hline
\end{tabular}
\caption{\label{demo-table}Environments overview.}
\end{center}
\end{table}

% \begin{tabular}{ | m{12 em} | m{4em} | m{5em} | m{4em} | m{4em} | m{4em} |} 

\begin{table}[!h]
\begin{center}
\begin{tabular}{  c  c  c  c  c  c } 
  \hline
    Hyper-parameter & BallCircle & BallGather & BallRun & BallReach & CartSafe \\
  \hline
    Batch-size & 64 & 64 & 64 & 64 & 64\\
    Epochs & 5 & 5 & 5 & 5 & 5\\
    Learning rate & 2e-4 & 2e-4 & 2e-4 & 2e-4 & 2e-4\\
    Optimizer & Adam & Adam & Adam & Adam & Adam\\
    Timesteps T & 32768 & 32768 & 32768 & 32768 & 32768\\
    Entropy co-efficient & 0.01 & 0.01 & 0.005& 0.01 & 0.001\\
    Clip range & 0.2 & 0.2 & 0.2 & 0.2 & 0.2\\
    GAE factor rewards $\lambda$ & 0.95 & 0.95 & 0.95 & 0.95  & 0.95\\
    Discount $\gamma$ &  0.99 & 0.99 & 0.99 & 0.99 & 0.99\\
    Safety discount $\gamma$ (\ref{section:V-C-ESTIMATE}) & 0.995 & 0.995 &  0.995 &  0.995 &  0.995\\
    Reward bias $b$ & 1.5 &  0.05 & 1 & 0.1 & 0\\
    $k$ ($V^{rc, \textbf{$k$}}$) & 2 & 4 & 4 & 4 & 5\\
    Cost factor $\beta$ & 0 & 15 & 0.5 & 0 & 3\\ 
\hline
\end{tabular}
\caption{\label{demo-table}Hyper-parameters used to train SCPO.}
\end{center}
\end{table}
\clearpage
\fi

\subsection{Lyapunov Approach}
\label{ap:Lyapunov-analysis}
% % In the early stages of this study, we considered basing our work on the Lyapunov approach \cite{lyapunov2002,lyapunov2018, 
% lyapunovcontinous}. After further investigation, we decided to explore a novel algorithm instead. 

In the following, we provide some theoretical properties and explain how the Lyapunov approach \cite{lyapunov2002,lyapunov2018, 
lyapunovcontinous} can be simplified in the deterministic case.   
% \subsubsection{Preliminaries}

We defined the generic bellman operator as:
\begin{equation*}
T_{\pi, h}[V](x)=\sum_a \pi(a, x)\left[h(x, a)+\sum_{x^{\prime} \in \mathcal{X}^{\prime}} P\left(x^{\prime}, x, a\right) V\left(x^{\prime}\right)\right]
\end{equation*}
We model the constraint reinforcement problem with a constraint Markov decision process (CMDP), which is defined by $\left(\mathcal{X}, \mathcal{A}, c, d, P, x_0, d_0\right)$. $\mathcal{X}$ and $\mathcal{A}$ are the state and action space, $d: \mathcal{X} \rightarrow \mathbb{R}_{\geq 0}$ is the immediate constraint cost and $d(x) \in\left[0, D_{\max }\right]$. $c: \mathcal{X}, \mathcal{A} \rightarrow \mathbb{R}$ is the cost function and $P(\cdot \mid x, a)$ is the transition probability.
\\
Let $\Delta(x)=\left\{\pi(\cdot \mid x): \mathcal{X} \rightarrow \mathbb{R}_{\geq 0 s}: \sum_a \pi(a \mid x)=1\right\}$ be the set of Markov stationary policies for any state $x\in \mathcal{X}$.
\\
The paper \cite{lyapunov2018} also defines $\mathrm{T}^*$ as a random variable corresponding to the first-hitting time of the terminal state $x_{\text {Term }}$ induced by policy $\pi$.
\\
We denote $\mathcal{C}_\pi(x) = \EX_\pi\left[ \sum_{t=0}^{T^*-1} c(x_i, a_i), x_0=x\right]$ and 
\\
$\mathcal{D}_\pi(x) = \EX_\pi\left[ \sum_{t=0}^{T^*-1} d(x_i), x_0=x\right]$.
\\
Given an initial state $x_0$ and a threshold $d_0$, We wish to solve the following problem denoted as $\OPT$:\\
\begin{equation*}
     \min _{\pi \in \Delta} \mathcal{C}_\pi\left(x_0\right) \quad \textrm{s.t} \ \mathcal{D}_\pi\left(x_0\right) \leq d_0
\end{equation*}
Let $\pi_B$ be a feasible policy of the $\OPT$ problem. The paper defines a non empty set of Lyapunov functions w.r.t state $x_0 \in \mathcal{X}$ and constraint threshold $d_0$ as:
\begin{equation*}
\begin{aligned}
&\mathcal{L}_{\pi_B}\left(x_0, d_0\right)=\left\{L: \mathcal{X} \rightarrow \mathbb{R}_{\geq 0}: T_{\pi_B, d}[L](x) 
\notag\right.
\\&
\phantom{=\;\;}
\left.
\leq L(x), \forall x \in \mathcal{X}^{\prime} ; L(x)=0, \forall x \in \mathcal{X} \backslash \mathcal{X}^{\prime} ; L\left(x_0\right) \leq d_0\right\}
\end{aligned}
\end{equation*}
\\
For any arbitrary Lyapunov function $L \in \mathcal{L}_{\pi_B}\left(x_0, d_0\right)$, denote by $\mathcal{F}_L(x)=$ $\left\{\pi(\cdot \mid x) \in \Delta: T_{\pi, d}[L](x) \leq L(x)\right\}$ the set of $L$-induced Markov stationary policies.
\\\\
{
\lemma{}
 There exists an auxiliary constraint cost $\epsilon: \mathcal{X}^{\prime} \rightarrow \mathbb{R}$ such that the Lyapunov function is given by $L_\epsilon(x)=\mathbb{E}\left[\sum_{t=0}^{\mathrm{T}^*-1} d\left(x_t\right)+\epsilon\left(x_t\right) \mid \pi_B, x\right], \forall x \in \mathcal{X}^{\prime}$, and $L_\epsilon(x)=0$ for $x \in \mathcal{X} \backslash \mathcal{X}^{\prime}$. Moreover, $L_\epsilon$ is equal to the constraint value function w.r.t. $\pi^*$, i.e., $L_\epsilon(x)=\mathcal{D}_{\pi^*}(x)$.
}
\\\\
Estimating $\epsilon$ requires knowledge of the optimal policy $\pi^\ast$. Therefore, the paper proposes to estimate $\epsilon^\ast(x) = \max_{x\in\mathcal{X}}(x) \geq 0$.
\\
For any arbitrary Lyapunov function $L \in \mathcal{L}_{\pi_B}\left(x_0, d_0\right)$, denote by
\\ $\mathcal{F}_L=$ $\left\{\pi(\cdot \mid x) \in \Delta: T_{\pi, d}[L](x) \leq L(x)\right\}$ the set of $L$-induced Markov stationary policies.
\\
The paper proposes an assumption that constrains the maximal distance between $\pi^\ast$ and $\pi_B$. Under it, they can guarantee that $\pi^\ast \in \mathcal{F}_{L_{\epsilon^\ast}}$  where
\begin{equation*}
    L_{\epsilon^\ast}(x)=\mathbb{E}\left[\sum_{t=0}^{\mathrm{T}^*-1} d\left(x_t\right)+\epsilon^\ast\left(x_t\right), x_0=x\right] \qquad  \text{and } \epsilon^\ast(x) \geq 0
\end{equation*}\\

\subsection{Lyapunov function properties}
{
\lemma{}
For an arbitrary policy $\pi$
\begin{equation*}
\begin{aligned}
&\forall{L}\in\mathcal{L}_{\pi}, \ \forall{k\in\mathbb{N}_{\geq 1}}:  \\& D_{\pi}(x) = \lim_{q\to\infty} T_{\pi,d}^{q}[L](x) \leq T_{\pi, d}^{k}[L](x) \leq  L(s)
\end{aligned}
\end{equation*}
}
\\
Proof:\\
For k=1, $T_{\pi, d}[L](x) \leq  L(s)$ follows by definition.\\
Assuming the property holds for k:\\
\begin{align*}
    T_{\pi, d}^{k+1}[L](x) &= d(x) + \sum_{a\in{\mathcal{A}}}\sum_{x^\prime\in\mathcal{X}} \pi(a, x) P(x^\prime, x, a) T_{\pi, d}^{k}[L](x)\\
    & \leq d(x) + \sum_{a\in{\mathcal{A}}}\sum_{x^\prime\in\mathcal{X}} \pi(a, x) P(x^\prime, x, a) L(x^\prime)\\
    & = L(x)
\end{align*}\\
$D_{\pi, d}(x) = \lim_{q\to\infty} T_{\pi,d}^{q}[L](x)$, because $T_{\pi, d}$ is a contraction mapping and $\mathcal{D}_\pi$ is a fix point.\\\\
A Lyapunov function is an upper bound on the expected cumulative cost.\\\\
We use the following notion: \\
$P_\pi[X_{t+1}=x_{t+1}, X_t=x_t] = \sum_{a\in\mathcal{A}}\pi(a, x_t) \ P[X_{t+1}=x_{t+1}, X_t=x_t, A_t=a]$
\\
{
\lemma{$\forall{x\in\mathcal{X}} \ \forall{T \geq 0}:$}
\begin{align}
\label{lemma:2}
    & \sum_{t=0}^{T} \sum_{x\in\mathcal{X}} P_\pi[X_t=x, X_0=x_0] d(x) \\& +  \sum_{x\in\mathcal{X}} P_\pi[X_{T+1}=x, X_0=x_0] \ L(x)\leq L(x_0) \nonumber
\end{align}
}
\\
Proof:\\
For $T=0$, the inequality holds by definition.\\
Suppose the inequality holds for arbitrary $T \geq 0$:
\begin{figure*}
\begin{align*}
    &\sum_{t=0}^{T} \sum_{x\in\mathcal{X}} P_\pi[X_t=x, X_0=x_0] d(x) +  \sum_{x\in\mathcal{X}} P_\pi[X_{T+1}=x, X_0=x_0] \ L(x)\leq L(x_0)\\
    &\Rightarrow 
    \begin{cases}
    \scalebox{0.85}{$\sum_{t=0}^{T} \sum_{x\in\mathcal{X}} P_\pi[X_t=x, X_0=x_0] d(x) +  \sum_{x\in\mathcal{X}} P_\pi[X_{T+1}=x, X_0=x_0] \ L(x)\leq L(x_0)$}\\
    L(x) = d(x) + \sum_{x^\prime\in\mathcal{X}} P[X_{T+2}=x^\prime, X_{T+1}=x] L(x^\prime)
    \end{cases}\\
    &\Rightarrow \scalebox{0.75}{ $\sum_{t=0}^{T+1} \sum_{x\in\mathcal{X}} P_\pi[X_t=x, X_0=x_0] d(x) + \sum_{x,x^\prime\in\mathcal{X}}P_\pi[X_{T+2}=x^\prime, X_{T+1}=x]P_\pi[X_{T+1}=x, X_0=x_0] L(x^\prime)$}\\
    &\Rightarrow \scalebox{0.85}{$\sum_{t=0}^{T+1} \sum_{x\in\mathcal{X}} P_\pi[X_t=x, X_0=x_0] d(x) + \sum_{x,x^\prime\in\mathcal{X}}P_\pi[X_{T+2}=x^\prime, X_{T+1}=x, X_0=x_0] L(x^\prime)$}\\
    &\Rightarrow \sum_{t=0}^{T+1} \sum_{x\in\mathcal{X}} P_\pi[X_t=x, X_0=x_0] d(x) + \sum_{x^\prime\in\mathcal{X}}P_\pi[X_{T+2}=x^\prime, X_0=x_0] L(x^\prime)
\end{align*}    
\end{figure*}

Therefore the inequality holds for all $T+1$.
\\\\
{
\corollary{For all $\pi$ in $\mathcal{F}_L$ The following inequality holds}:
\begin{equation}
\label{corollary:L_d0}
    \forall{x\in\mathcal{X}}: \qquad \max_{t\geq0} P_\pi[X_t=x, X_0=x_0] L(x) \leq d_0
\end{equation}
}
\\
% Proof:\\

\begin{proof}
 Let $x^\prime \in \mathcal{X}$ and $T\geq0$
 % \begin{figure*}
 \begin{align*}
    &\begin{cases}    
        \sum_{t=0}^{T} \sum_{x\in\mathcal{X}} P_\pi[X_t=x, X_0=x_0] d(x) + \\  \sum_{x\in\mathcal{X}} P_\pi[X_{T+1}=x, X_0=x_0] \ L(x)\leq L(x_0)
        \\
        \\
        -\sum_{t=0}^{T} \sum_{x\in\mathcal{X}} P_\pi[X_t=x, X_0=x_0] d(x)\leq 0
    \end{cases}\\
    &\Rightarrow \sum_{x\in\mathcal{X}} P_\pi[X_{T+1}=x, X_0=x_0] \ L(x)\leq L(x_0)\\
    &\Rightarrow P_\pi[X_{T+1}=x^\prime, X_0=x_0] \ L(x^\prime)\leq L(x_0)\\
\end{align*}    
 % \end{figure*}

Therefore:\\
\begin{align*}
    &\begin{cases}
    \forall{t\geq1}: \quad P_\pi[X_t=x^\prime, X_0=x_0] \ L(x^\prime)\leq L(x_0)\\
    P_\pi[X_0=x^\prime, X_0=x_0] \ L(x^\prime) = \mathbf{1}[x_0 = x^\prime] L(x_0)
    \end{cases}\\
    &\Rightarrow\begin{cases}
    \forall{t\geq1}: \quad P_\pi[X_t=x^\prime, X_0=x_0] \ L(x^\prime)\leq L(x_0)\\
    P_\pi[X_0=x^\prime, X_0=x_0] \ L(x^\prime) \leq L(x_0)
    \end{cases}\\
     &\Rightarrow \forall{t\geq0} \ P_\pi[X_t=x^\prime, X_0=x_0] \ L(x^\prime)\leq L(x_0)\\
     &\Rightarrow \max_{t\geq0} \ P_\pi[X_t=x^\prime, X_0=x_0] \ L(x^\prime)\leq L(x_0)\\
     &\Rightarrow \max_{t\geq0} \ P_\pi[X_t=x^\prime, X_0=x_0] \ L(x^\prime)\leq d_0\\
\end{align*}   
\end{proof}

% \subsubsection{Deterministic case}
% In this section, 

\textbf{Deterministic cases}: we assume a deterministic environment. Therefore, an optimal deterministic policy exists $\pi^\ast$ for the $\OPT$ problem.\\
Let $\tau^\ast$ be the optimal trajectory induced by $\pi^\ast$ of length $T^\ast$. For all $t\in [0, T^\ast-1]$, we defined $x_t^\ast \in \mathcal{X}$ such that $P_\pi^\ast[X_t = x_t^\ast, X_0=x_0] = 1$. Hence, $\tau^\ast = x_0 x_1^\ast, x_2^\ast ... x_{T-1}^\ast$\\
Let $L(x)=D_{\pi_B}(x)$ for all $x\in\mathcal{X}$. 
\\
{
\lemma{}{}
\begin{equation}
\label{lemma:L_SMALLER_THAN_D0}
    \forall {i\in[0, T^\ast-1]} \quad L(x_i^\ast)\leq d_0  \quad \text{and} \quad T_{\pi^\ast, d}[L](x_i^\ast) \leq d_0
\end{equation}
}
Proof:\\
Let $t \in[0, T^{\ast}-1]$ arbitrary. $\LEPSILONESTAR$ is a valid Lyapunov function for $\pi^\ast$ Because $\pi^\ast \in \mathcal{F}_{\LEPSILONESTAR}$
\begin{align*}
    &\forall{x\in\mathcal{X}} \ \max_{t^\prime\geq0} P_{\pi^\ast}[X_{t^\prime}=x, X_0=x_0] \LEPSILONESTAR(x) \leq d_0 \\&\Rightarrow \max_{t^\prime\geq0} P_{\pi^\ast}[X_{t^\prime}=x_t^\ast, X_0=x_0] \LEPSILONESTAR(x_t^\ast) \leq d_0\\
    & \Rightarrow P_{\pi^\ast}[X_{t}=x_t^\ast, X_0=x_0] \LEPSILONESTAR(x_t^\ast) \leq d_0\\
    & \Rightarrow \LEPSILONESTAR(x_t^\ast) \leq d_0\\
    & \Rightarrow \begin{cases}
        L(x_t^\ast) \leq d_0\\
        T_{\pi^\ast, d}[\LEPSILONESTAR](x_i^\ast) \leq d_0
    \end{cases}\\
    & \Rightarrow \begin{cases}
        L(x_t^\ast) \leq d_0\\
        T_{\pi^\ast, d}[L](x_i^\ast) \leq d_0
    \end{cases}
\end{align*}
\\\\
$\forall{a\in \mathcal{A}}$, we define $\pi_1^\ast \in \Delta$ as follows:\\
\begin{equation*}
    \pi_1^\ast(a, x) = \begin{cases}
    \pi^\ast(a, x) \quad \text{if} \ \exists {t \geq 0} \ P_{\pi^\ast}[X_t=x, X_0=x_0] > 0\\
    \pi_B(a, x) \quad \text{otherwise}.\\
    \end{cases}
\end{equation*}
$\pi_1^\ast$ is also optimal because it agrees with $\pi^\ast$ for all states on the optimal trajectory $\tau^\ast$.
\\
{
\corollary{}{}
Let $L(x) = D_{\pi_B}(x)$ and assume $\pi^\ast\in\mathcal{F}_{\LEPSILONESTAR}$
\begin{equation}
    \label{Corollary:L_induces_pie_star}
   \pi_1^\ast \in \left\{\pi(., x), \forall{x\in\mathcal{X}} \quad T_{\pi, d}[L](x) \leq \max(d_0, L(x)) \right\}
\end{equation}
}
\\
% Proof:\\

\begin{proof}
For all states $x$ such that $P_{\pi^\ast}[X_t=x, X_0=x_0] \geq 0$, $\pi^\ast = \pi_1^\ast$ and the inequality holds directly from \eqref{lemma:L_SMALLER_THAN_D0}.\\
For other states $\pi_B$ and $\pi_1^\ast$ are equal, the inequality holds because $L$ is a valid Lyapunov function with respect to $\pi_B$.
\\\\
Hence, instead of estimating $\epsilon^\ast: \mathcal{X} \rightarrow \mathbb{R}_{\geq 0 s}$ and searching for the optimal policy in the set $\mathcal{F}_{\LEPSILONESTAR}$ we can instead search for it in the above-mentioned set, which does not require extra computations.\\    
\end{proof}

\section{Proofs}

{
\label{proof:vc-diff-Q}
Theorem \ref{theorem:v_c_diff_Q} proof:
\begin{align*}
    \nabla V^c_\pi(s) &= \sum_{a\in\mathcal{A}} \nabla(\pi(a\kgiven s) \ Q^c_\pi(s, a))\\
    &= \sum_{a\in\mathcal{A}} \nabla \pi(a\kgiven s) \ Q^c_\pi(s, a) + \sum_{a\in\mathcal{A}} \pi(a\kgiven s) \ \nabla Q^c_\pi(s, a)\\
    &=\scalebox{0.75}{$\sum_{a\in\mathcal{A}} \pi(a\kgiven s) \nabla \log(\pi(a\kgiven s)) \ Q^c_\pi(s, a) + \sum_{a\in\mathcal{A}} \pi(a\kgiven s) \ \sum_{s^\prime} p(s^\prime\kgiven s, a) \ \nabla V^c_\pi(s^\prime)$} \quad  \eqref{Q_V_C}\\
    &= \EX_\pi\left[\sum_{t=0} \nabla\log(\pi(S_t\kgiven A_t)) Q_\pi^c(S_t, A_t) \kgiven S_0=s \right].
\end{align*}
}
%--------------------------------------------
{
\label{proof:v-c-diff-A}
Corollary \ref{corollary:v_c_diff_A} proof:\\
\begin{align*}
\label{proof:v_c_diff_A}
    & \EX_\pi\left[\sum_{t=0} \nabla\log(\pi(S_t, A_t)) Q_\pi^c(S_t, A_t) \kgiven S_0=s \right] \\
    &= \sum_{t=0}\sum_{s^\prime\in\mathcal{S}} P_\pi[S_t=s^\prime \kgiven S_0=s] \sum_{a\in\mathcal{A}} \pi(a\kgiven s^\prime) \frac{\nabla\pi(a\kgiven s)}{\pi(a\kgiven s)} A^c_\pi(s, a)\\
    &= \sum_{t=0}\sum_{s^\prime\in\mathcal{S}} P_\pi[S_t=s^\prime \kgiven S_0=s] \sum_{a\in\mathcal{A}} \pi(a\kgiven s^\prime) \frac{\nabla\pi(a\kgiven s)}{\pi(a\kgiven s)} Q^c_\pi(s^\prime, a) - V(s^\prime)\\
    &= \sum_{t=0}\sum_{s^\prime\in\mathcal{S}} P_\pi[S_t=s^\prime \kgiven S_0=s] \sum_{a\in\mathcal{A}} \pi(a\kgiven s^\prime) \frac{\nabla\pi(a\kgiven s)}{\pi(a\kgiven s)} Q^c_\pi(s^\prime, a) -  \sum_{a\in\mathcal{A}} \nabla \pi(a\kgiven s^\prime)V(s^\prime)\\
    &= \sum_{t=0}\sum_{s^\prime\in\mathcal{S}} P_\pi[S_t=s^\prime \kgiven S_0=s] \sum_{a\in\mathcal{A}} \pi(a\kgiven s^\prime) \frac{\nabla\pi(a\kgiven s)}{\pi(a\kgiven s)} Q^c_\pi(s^\prime, a) - \nabla 1 \ V(s^\prime)\\
    &= \sum_{t=0}\sum_{s^\prime\in\mathcal{S}} P_\pi[S_t=s^\prime \kgiven S_0=s] \sum_{a\in\mathcal{A}} \pi(a\kgiven s^\prime) \frac{\nabla\pi(a\kgiven s)}{\pi(a\kgiven s)} Q^c_\pi(s^\prime, a)\\
    &= V^c_\pi(s).\\
\end{align*}
}

%--------------------------------------------
{
\label{proof:trust-region-1}
Theorem \ref{thorem:trust-region-1} proof:\\
\begin{align*}
    &(V_{\pi^\prime}^c - V_{\pi}^c)(s) \\&= \sum_{a\in \mathcal{A}} \pi^\prime(a\kgiven s) \ Q_{\piprime}^c(s, a) - \sum_{a\in \mathcal{A}} \pi(a\kgiven s) \ Q_{\pi}^c(s, a)\\
    &= \sum_{a\in \mathcal{A}} \pi^\prime(a\kgiven s) \ Q_{\piprime}^c(s, a) - \sum_{a\in \mathcal{A}} \pi(a\kgiven s) \ Q_{\pi}^c(s, a) \pm \sum_{a\in \mathbf{A}} \piprime(a\kgiven s) \ Q_{\pi}^c(s, a)\\
    &= \sum_{a\in \mathcal{A}} \pi^\prime(a\kgiven s) \ (Q_{\piprime}^c(s, a) - Q_\pi^c(s, a)) + \sum_{a\in \mathcal{A}} (\piprime(a\kgiven s) - \pi(a\kgiven s)) \ Q_{\pi}^c(s, a)\\
    &= \sum_{a\in \mathcal{A}}\sum_{s^\prime \in \mathcal{S}} \pi^\prime(a\kgiven s) \ p(s^\prime\kgiven s, a) \ (V_{\piprime}^c- V_\pi^c)(s^\prime) + \sum_{a\in \mathcal{A}} \piprime(s, a)(1 - \frac{\pi(a\kgiven s)}{\piprime(a\kgiven s)}) \ Q_{\pi}^c(s, a) \quad \\
    &= \EX_{\pi^\prime}\left[\sum_t (1-\frac{\pi(A_t\kgiven S_t)}{\pi^\prime(A_t\kgiven S_t)}) Q_\pi^c(S_t, A_t) \kgiven S_0=s\right], \ \eqref{Q_V_C}.
\end{align*}
}
%-----------------------------------------------
{
\definition{}
$(\pi^\prime, \pi)$ is an $\alpha$-coupled policy pair if it defines a joint distribution $P(a, a^\prime \kgiven s)$, such that $P(a\neq a^\prime \kgiven s) \leq \alpha$ for all $s \in \mathcal{S}$. $\pi$ and $\pi^\prime$ will denote the marginal distributions of $a$ and $a^\prime$ respectively.
}

%--------------------------------------------
{
\label{proof:trust-region-2}
Corollary \ref{corollary:trust-region-2} proof:\\
\begin{align*}
    & \EX_{\pi^\prime}\left[\sum_{t=0} \gamma^t (1-\frac{\pi(A_t\kgiven S_t)}{\pi^\prime(A_t\kgiven S_t)}) A_\pi(S_t, A_t) \kgiven S_0=s\right]\\
    &= \sum_{t=0}\sum_{s^\prime\in\mathcal{S}} P_\pi[S_t=s^\prime \kgiven S_0=s] \gamma^t \sum_{a\in\mathcal{A}}   (\piprime(a\kgiven s) -\pi(a\kgiven s)) A_\pi(s^\prime, a)\\
   &= \scalebox{0.75}{$\sum_{t=0}\sum_{s^\prime\in\mathcal{S}} P_\pi[S_t=s^\prime \kgiven S_0=s] \gamma^t \sum_{a\in\mathcal{A}}   (\piprime(a\kgiven s) -\pi(a\kgiven s)) Q_\pi(s^\prime, a) - V_\pi(s^\prime) \ \sum_{a\in \mathcal{A}}\piprime(a\kgiven s) -\pi(a\kgiven s)$} \\
   &= \sum_{t=0}\sum_{s^\prime\in\mathcal{S}} P_\pi[S_t=s^\prime \kgiven S_0=s] \gamma^t \sum_{a\in\mathcal{A}}   (\piprime(a\kgiven s) -\pi(a\kgiven s)) Q_\pi(s^\prime, a) - V_\pi(s^\prime) \ 0\\
    &= \EX_{\pi^\prime}\left[\sum_{t=0} \gamma^t (1-\frac{\pi(A_t\kgiven S_t)}{\pi^\prime(A_t\kgiven S_t)}) Q_\pi(S_t, A_t) \kgiven S_0=s\right].
\end{align*}
}

%--------------------------------------------
{
\label{proof:trust-region-estimator-1}
Theorem \ref{thorem:trust-region-estimator-1} proof:\\
We define $H(s) = \EX_\pi\left[(\frac{\pi^\prime(A \kgiven S)}{\pi(A\kgiven S)} -1) \ A_\pi(S, A) \kgiven S=s \right]$.\\
Let $s\in\mathcal{S}$. The following equality holds:
\begin{equation}
\EX_{\pi^\prime}\left[\sum_{t=0} \gamma^t (1-\frac{\pi(A_t\kgiven S_t)}{\pi^\prime(A_t\kgiven S_t)}) A_\pi(S_t, A_t) \kgiven S_0=s\right] = \EX_{\pi^\prime}\left[\sum_{t=0} \gamma^t H(s) \kgiven S_0=s\right].
\end{equation}
Proof:
\begin{align*}
    &\EX_{\pi^\prime}\left[\sum_{t=0} \gamma^t (1-\frac{\pi(A_t\kgiven S_t)}{\pi^\prime(A_t\kgiven S_t)}) A_\pi(S_t, A_t) \kgiven S_0=s\right]\\
    &= \sum_{t=0}\sum_{s^\prime\in\mathcal{S}} P_\pi[S_t=s^\prime \kgiven S_0=s] \gamma^t \sum_{a\in\mathcal{A}}   (\piprime(a\kgiven s) -\pi(a\kgiven s)) A_\pi(s^\prime, a)\\
    &= \sum_{t=0}\sum_{s^\prime\in\mathcal{S}} P_\pi[S_t=s^\prime \kgiven S_0=s] \gamma^t \sum_{a\in\mathcal{A}}   \pi(a\kgiven s) (\frac{\piprime(a\kgiven s)}{\pi(a\kgiven s)} -1) A_\pi(s^\prime, a)\\
    &= \sum_{t=0}\sum_{s^\prime\in\mathcal{S}} P_\pi[S_t=s^\prime \kgiven S_0=s] \gamma^t \EX_\pi\left[(\frac{\pi^\prime(a \kgiven s)}{\pi(a\kgiven s)} -1) \ A_\pi(S, A) \kgiven S=s^\prime \right] \\
    &= \EX_{\pi^\prime}\left[\sum_{t=0} \gamma^t H(S_t) \kgiven S_0=s\right].
\end{align*}
Instead of sampling trajectories using $\pi^\prime$ and $\pi$ independently, we sample a pair of trajectories at the same time using the $\alpha$ coupled policies $(\pi^\prime, \pi)$:
\begin{equation*}
    \EX_{\pi^\prime}\left[\sum_{t=0} \gamma^t H(S_t)\right] - \EX_{\pi}\left[\sum_{t=0} \gamma^t H(S_t)\right] = \EX_{(\pi^\prime, \pi)}\left[\sum_{t=0} \gamma^t (H(S_t^\prime) - H(S_t))\right].
\end{equation*}
Because the trajectories can diverge, let $S_t^\prime$ and $S_t$ respectively denote the states at time $T$ of the first and second trajectory. The first trajectory is generated using $\pi^\prime$, while the second trajectory is generated using $\pi$.
\begin{equation*}
    P_{(\pi^\prime, \pi)}[S_t^\prime = S_t] \leq (1-\alpha)^t \qquad P_{(\pi^\prime, \pi)}[S_t^\prime \neq S_t] \geq 1-(1-\alpha)^t.
\end{equation*}
For two trajectories to agree on a state $s_t$, they must either agree on all actions up to time $t$ with probability $(1-\alpha)^t$, or they can diverge at some point but still end up at the same state $s_t$.
\begin{align*}
    &\EX_{(\pi^\prime, \pi)}\left[\gamma^t (H(S_t^\prime) - H(S_t))\right] \\&= \EX_{(\pi^\prime, \pi)}\left[\gamma^t (H(S_t^\prime) - H(S_t)) \kgiven S_t^\prime=S_t\right] \ P[S_t^\prime = S_t] \\
     & \qquad\qquad + \EX_{(\pi^\prime, \pi)}\left[\gamma^t (H(S_t^\prime) - H(S_t)) \kgiven S_t^\prime \neq S_t\right] \ P[S_t^\prime \neq S_t]\\
    &= \EX_{(\pi^\prime, \pi)}\left[\gamma^t (H(S_t^\prime) - H(S_t)) \kgiven S_t^\prime \neq S_t\right] \ P[S_t^\prime \neq S_t]\\
    &\geq \EX_{(\pi^\prime, \pi)}\left[\gamma^t (H(S_t^\prime) - H(S_t)) \kgiven S_t^\prime \neq S_t \right] (1 -(1-\alpha)^t)\\
    & \geq -2\gamma^t \ \max_{s,a} \lvert(\frac{\pi^\prime(a\kgiven s)}{\pi(a\kgiven s)} - 1) A(s, a)\rvert \ (1 -(1-\alpha)^t)\\
    &= -2\gamma^t \epsilon \ (1 -(1-\alpha)^t).
\end{align*}
Therefore, we can conclude:
\begin{align*}
\EX_{\pi^\prime}\left[\sum_{t=0} \gamma^t H(S_t)\right] - \EX_{\pi}\left[\sum_{t=0} \gamma^t H(S_t)\right] &= \EX_{(\pi^\prime, \pi)}\left[\sum_{t=0}\gamma^t (H(S_t^\prime) - H(S_t))\right]\\ 
    &= \sum_{t=0}\EX_{(\pi^\prime, \pi)}\left[\gamma^t (H(S_t^\prime) - H(S_t))\right]\\
    &\geq \sum_{t=0} -2\gamma^t \epsilon \ (1 -(1-\alpha)^t)\\
    &\geq -2\gamma \epsilon \frac{1}{1-\gamma} \ \frac{1}{1 -\gamma (1-\alpha)}\\
    &= -2\gamma \epsilon \ \frac{\alpha \gamma}{(1-\gamma)(1 -\gamma (1-\alpha))}\\
    &\geq  -\frac{2\gamma^2 \alpha \ \epsilon}{(1-\gamma)^2}.
\end{align*}
Using the same arguments from \cite{trpo}, if $\max_s D_{TV}(\pi^\prime(a\kgiven .), \pi(a\kgiven .)) \leq \alpha$, we can define an $\alpha$-coupled policy pair $(\pi^\prime, \pi)$.
}

%--------------------------------------------
{
\label{proof:trust-region-estimator-2}
Theorem \ref{theorem:trust-region-estimator-2} proof:\\
For notational convenience, we define $H^\prime(s, a) = (1 - \frac{\pi(a\kgiven s)}{\pi^\prime(a\kgiven s)}) A_\pi(s, a)$.\\
We use the same principle and notation from the previous proof. In addition, let $A_t^\prime$ and $A_t$ be the action chosen by the $\alpha$-coupled policies $(\pi^\prime, \pi)$ at timestep $t$. We denote the event $D$ by \{$S_t^\prime \neq S_t \lor A_t^\prime \neq A_t$\}\\
\begin{align*}
    \begin{cases}
        P_{(\pi^\prime, \pi)}[S_t^\prime = S_t] \leq (1-\alpha)^t \\
        P_{(\pi^\prime, \pi)}[A_t^\prime = A_t \kgiven S_t^\prime = S_t] = 1-\alpha 
    \end{cases}
        &\Rightarrow P_{(\pi^\prime, \pi)}[S_t^\prime = S_t, A_t^\prime = A_t] \leq (1-\alpha)^{t+1}\\
        &\Rightarrow P_{(\pi^\prime, \pi)}[S_t^\prime \neq S_t 	\lor A_t^\prime \neq A_t] \geq 1 -(1-\alpha)^{t+1}\\
        &\Rightarrow P_{(\pi^\prime, \pi)}[D] \geq 1 -(1-\alpha)^{t+1}.
\end{align*}

If both policies agree on the state and action at time t, the expected value is $0$:
\begin{equation*}
        \EX_{(\pi^\prime, \pi)}\left[\gamma^t (H^\prime(S_t^\prime, A_t^\prime) - H^\prime(S_t, A_t)) \kgiven  S_t^\prime = S_t, A_t^\prime = A_t \right] = 0.
\end{equation*}
Otherwise:
\begin{align*}
    \EX_{(\pi^\prime, \pi)}\left[\gamma^t (H^\prime(S_t^\prime, A_t^\prime) - H^\prime(S_t, A_t)) \kgiven  D \right] \ P_{(\pi^\prime, \pi)}[D]&\geq - 2 \gamma^t \epsilon^\prime (1-(1- \alpha)^{t+1}).
\end{align*}
Combining the equations, we deduce the following:\\
\begin{align*}
    &\EX_{\pi^\prime}\left[\sum_{t=0} \gamma^t H^\prime(S_t, A_t)\right] - \EX_{\pi}\left[\sum_{t=0} \gamma^t H^\prime(S_t, A_t)\right] \\&= \EX_{(\pi^\prime, \pi)}\left[\sum_{t=0}\gamma^t  H^\prime(S_t^\prime, A_t^\prime) -  H^\prime(S_t, A_t) \right]\\
    &= \sum_{t=0} \EX_{(\pi^\prime, \pi)}\left[\gamma^t  H^\prime(S_t^\prime, A_t^\prime) -  H^\prime(S_t, A_t) \right]\\
    &\geq \sum_{t=0} -2\gamma^t \epsilon^\prime \ (1 -(1-\alpha)^t{+1})\\
    &\geq -2\gamma \epsilon^\prime \frac{1}{1-\gamma} \ \frac{1}{1 -\gamma (1-\alpha)}\\
    &= -2\gamma \epsilon^\prime \ \frac{\alpha \gamma}{(1-\gamma)(1 -\gamma (1-\alpha))}\\
    &\geq  -\frac{2\gamma^2 \alpha \ \epsilon^\prime}{(1-\gamma)^2}.
\end{align*}
}

%--------------------------------------
{
Equation \ref{V_R} proof:
\label{proof:V_R}
\begin{align*}
    &V_\pi^r(s) \\&= \EX_\pi\left[\sum_{t=0} \gamma^t r(S_t, A_t) \ \prod_{t=0} f(S_t) \kgiven S_0=s\right]\\
    &= \sum_{a\in\mathcal{A}} \pi(a\kgiven s) E_\pi\left[r(s, a) \prod_{t=0} f(S_t) \kgiven S_0=s, A_0=a\right] \\
    &\qquad\qquad+ \EX_\pi\left[\sum_{t=1} \gamma^t r(S_t, A_t) \prod_{t=0} f(S_t) \kgiven S_0=s, A_0=a\right]\\ 
    &= \scalebox{0.85}{$\sum_{a\in\mathcal{A}} \pi(a \kgiven s) \ r(s, a) \ Q^c_\pi(s, a) + \sum_{s^\prime \in \mathcal{S}} p(s^\prime \kgiven s, a) f(s) \EX_\pi\left[\sum_{t=1} \gamma^t r(S_t, A_t) \prod_{t=1} f(S_t) \kgiven S_1=s^\prime\right]$}\\ 
    &= \scalebox{0.85}{$\sum_{a\in\mathcal{A}} \pi(a \kgiven s) \ r(s, a) \ Q^c_\pi(s, a) + \sum_{s^\prime \in \mathcal{S}} p(s^\prime \kgiven s, a) \EX_\pi\left[\sum_{t=1} \gamma^t r(S_t, A_t) \prod_{t=1} f(S_t) \kgiven S_1=s^\prime\right]$} \\
    &= \scalebox{0.85}{$\sum_{a\in\mathcal{A}} \pi(a \kgiven s) \ r(s, a) \ Q^c_\pi(s, a) + \sum_{s^\prime \in \mathcal{S}} p(s^\prime \kgiven s, a) \gamma \EX_\pi\left[\sum_{t=0} \gamma^t r(S_t, A_t) \prod_{t=0} f(S_t) \kgiven S_0=s^\prime\right]$}\\
    &= \sum_{a\in\mathcal{A}} \pi(a \kgiven s) \ r(s, a) \ Q^c_\pi(s, a) + \sum_{s^\prime \in \mathcal{S}} p(s^\prime \kgiven s, a) \gamma  V^r_\pi(s^\prime)\\
    &= \EX_\pi\left[\sum_{t=0} \gamma^t r(S_t, A_t) \ Q^c_\pi(S_t, A_t) \kgiven S_0=s \right].
\end{align*}
}

%-------------------------
{
\label{proof:V-rc-bar}
Theorem \ref{V-rc-bar} proof:
\begin{align*}
    &(\bar{V}_{\pi^\prime}^{r,k} - V_\pi^{r,k})(s) \\&= \sum_{a\in \mathcal{A}} \pi^\prime(a\kgiven s) \ \bar{Q}_{\piprime}^{r,k}(s, a) - \sum_{a\in \mathcal{A}} \pi(a\kgiven s) \ Q_{\pi}^{r,k}(s, a)\\
    &= \sum_{a\in \mathcal{A}} \pi^\prime(a\kgiven s) \ \bar{Q}_{\piprime}^{r,k}(s, a) - \sum_{a\in \mathcal{A}} \pi(a\kgiven s) \ Q_{\pi}^{r,k}(s, a) \pm \sum_{a\in \mathbf{A}} \piprime(a\kgiven s) \ \bar{Q}_{\pi}^{r,k}(s, a)\\
    &= \sum_{a\in \mathcal{A}} \pi^\prime(a\kgiven s) \ (\bar{Q}_{\piprime}^{r,k}(s, a) - Q_\pi^{r,k}(s, a)) + \sum_{a\in \mathcal{A}} (\piprime(a\kgiven s) - \pi(a\kgiven s)) \ Q_{\pi}^{r,k}(s, a)\\
    &= \sum_{a\in \mathcal{A}}\sum_{s^\prime \in \mathcal{S}} \pi^\prime(a\kgiven s) p(s^\prime\kgiven s, a) \gamma (\bar{V}_{\piprime}^{r,k} - V_\pi^{r,k})(s^\prime) + \sum_{a\in \mathcal{A}} \piprime(s, a)(1 - \frac{\pi(a\kgiven s)}{\piprime(a\kgiven s)}) 
 Q_{\pi}^{r,k}(s, a) \quad\\
    &= \EX_{\pi^\prime}\left[\sum_t \gamma^t (1-\frac{\pi(A_t\kgiven S_t)}{\pi^\prime(A_t\kgiven S_t)}) Q_\pi^{r,k}(S_t, A_t) \kgiven S_0=s\right].
\end{align*}
}
%------------------------------
{
\theorem{}
\begin{equation}
\label{Q_V_C}
    Q^c_\pi(s, a) = \sum_{s^\prime \in \mathcal{X}} p(s^\prime \kgiven s, a) \  V^c_\pi(s^\prime).
\end{equation}
}
Proof:
\begin{align*}
   Q^c_\pi(s, a) &= \EX_\pi\left[\prod_{t=0} f(S_t) \kgiven S_0=s, A_0=a \right]\\
    &= \sum_{s^\prime\in\mathcal{S}} p(s^\prime \kgiven s, a) \EX_\pi\left[\prod_{t=0} f(S_t) \kgiven S_0=s, A_t=a, S_1=s^\prime \right] \\
    &= \sum_{s^\prime\in\mathcal{S}} p(s^\prime \kgiven s, a) f(s) \ \EX_\pi\left[\prod_{t=1} f(S_t) \kgiven S_0=s, A_t=a, S_1=s^\prime \right]\\
    &= \sum_{s^\prime\in\mathcal{S}} p(s^\prime \kgiven s, a) \ f(s) \ \EX_\pi\left[\prod_{t=0} f(S_t) \kgiven S_0=s^\prime \right]\\
    &= \sum_{a\in\mathcal{A}} \pi(a\kgiven s) \sum_{s^\prime\in\mathcal{S}} p(s^\prime \kgiven s, a) \ V^c_\pi(s^\prime).
\end{align*}

The cumulative cost can only increase when transitioning from one state to another. If $f(s_t) = 0$ then all consecutive states are also unsafe. Therefore, the following holds: \\
$ f(s) p(s^\prime\kgiven s, a) \EX_\pi\left[\prod_{t=1}f(S_t) \kgiven S_1=s^\prime \right] = p(s^\prime\kgiven s, a) \EX_\pi\left[\prod_{t=1} f(S_t) \kgiven S_1=s^\prime \right] = 0$. The equality is trivial when $f(s) = 1$.
\section{$Q^c$ and $V^c$ estimate}
\label{appendix-section:V-C-ESTIMATE}
Let $\tau = (s_0, a_0, s_1, a_1 ... a_{T-1}, s_{T-1})$ be a trajectory. $Q^c_\pi(s_t, a_t)$ can be expressed as follows:
\begin{align*}
    Q^{c, (1)}_\pi(s_t, a_t) &= \EX_\pi[f(S_t) \ V^c_\pi(S_{t+1}) \kgiven S_t=s_t, A_t=a_t],\\
    Q^{c, (2)}_\pi(s_t, a_t) &= \EX_\pi[f(S_t) \ f(S_{t+1}) \ V^c_\pi(S_{t+1}) \kgiven S_t=s_t, A_t=a_t],\\
    & ...\\
    Q^{c, (k)}_\pi(s_t, a_t) &= \EX_\pi[f(S_t) f(S_{t+1}) ... f(S_{t+k-1}) \ V^c_\pi(S_{t+k})) \kgiven S_t=s_t, A_t=a_t].
\end{align*}

Let $\hat{Q}^{c, (k)}_\pi$ be the one sample estimate of $Q^{c, (k)}_\pi$:\\
\begin{align*}
    \hat{Q}^{c, (1)}_\pi(s_t, a_t) &= f(s_t) \ V^c_\pi(s_{t+1})\\
    \hat{Q}^{c, (2)}_\pi(s_t, a_t) &= f(s_t) \ f(s_{t+1}) \ V^c_\pi(s_{t+2}) = f(s_{t+1}) \ V^c_\pi(s_{t+2})\\
    & ...\\
    \hat{Q}^{c, (k)}_\pi(s_t, a_t) &= f(s_t) \ f(s_{t+1}) ... f(s_{t+k-1}) \ V^c_\pi(s_{t+k}) = f(s_{t+k-1}) \ V^c_\pi(s_{t+k}).
\end{align*}
\\
Any weighted average of $\hat{Q}^{c, (k)}_\pi$ can be used as an estimate. Let $\Bar{Q}{^c_\pi}$ be such an estimate. We define the estimate of the safety advantage as follows:
\begin{equation*}
    \hat{A}^c_\pi(s, a) = \Bar{Q}^c_\pi(s, a) - V^c_\pi(s).
\end{equation*}
\\
We can also introduce the notion of discount when estimating $V^c$: Let $T$ be the episode length, we define $\hat{V}^c_\gamma$ as follows:
\begin{equation}
    \forall{\gamma \in (0, 1]}, \quad \hat{V}^c_\gamma(s_t) = 
    \begin{cases}
    (1-\gamma) \ f(s_t) + \gamma \hat{V}^c_\gamma(s_{t+1}),  \qquad &\text{if} \ \ 0\leq t < T-1;\\
   f(s_{T-1}), \qquad &\text{otherwise}.
    \end{cases}
\end{equation}
We notice that $V^c_1(s_t) = f(S_{T-1})$, which is the one sample estimate of $V^c(s_t)$.\\
If $f(S_{T-1}) = 1$, then for all $ \gamma \in (0, 1], \ \ V^c_\gamma(s_t) = 1$.\\
\begin{figure}[!ht]
  \centering
  \begin{minipage}[b]{0.495\textwidth}
    \includegraphics[width=\textwidth]{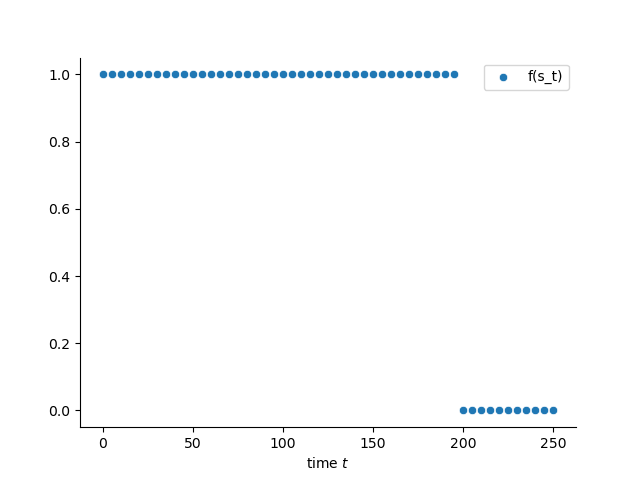}
  \end{minipage}
  \hfill
  \begin{minipage}[b]{0.495\textwidth}
    \includegraphics[width=\textwidth]{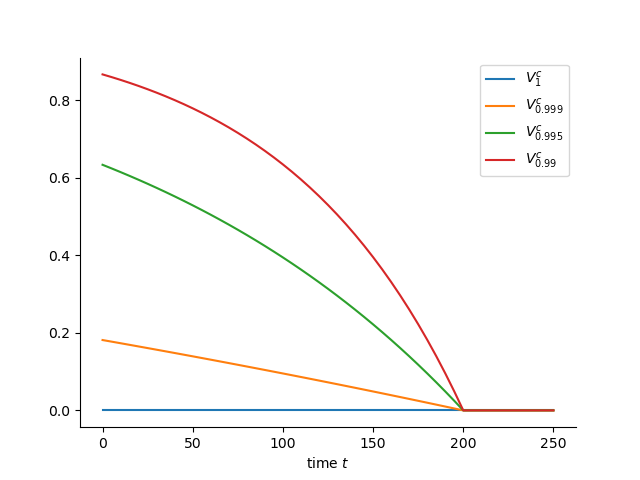}
  \end{minipage}
      \caption{The left graph represents the value of $f(s_t)$ for a specific episode. In the right graph we plot $V^c_\gamma$ for different values of $\gamma$}
\end{figure}

Unrolling the recursion, $V^c_\gamma$ can be written as follows:
\begin{equation*}
    V^c_\gamma(s_t) = \begin{cases}
    \sum_{k=0}^{T-t-1} (1-\gamma) \gamma^k f(s_{t+k}), \qquad &\text{if} \ \ 0 \leq t < T-1;\\
    f(S_{T-1}), \qquad &\text{otherwise}.
    \end{cases} 
\end{equation*}

Choosing an appropriate $\gamma$ value can reduce variance when estimating $V^c$. If the current state is unsafe, depending on the problem at hand, we can roughly analyze how much previous states led to unsafely.
The upper bound of $V^c_\gamma$ is $1$. Therefore, it can still be interpreted as the probability of visiting an unsafe state.
\begin{align*}
    V^c_\gamma(s_t) &= \sum_{k=0}^{T-t-1} (1-\gamma) \gamma^k f(s_{t+k})\\
    &\leq \sum_{k=0}^{T-t-1} (1-\gamma) \gamma^k\\
    &\leq (1-\gamma) \frac{1-\gamma^{T-t}}{1-\gamma}\\
    &\leq 1-\gamma^{T-t}\\
    &\leq 1.
\end{align*}

\begin{comment}
\section{Trust region} \label{appendix-chapter:trust_region}
To find an optimal policy, we randomly initialize a policy $\pi$, then iteratively improve it by collecting trajectories and following the gradient of the objective function. When iterating, stochastic gradient ascent is used with batches of configurable size. After one epoch, the policy changes from $\pi$ to $\pi^\prime$. The trajectories were collected using $\pi$, which is different from the intermediate policy we generated after one epoch. TRPO~\cite{trpo} addresses this problem and proposes a solution that guarantees the objective function's improvement using samples generated by the initial policy $\pi$. In this section, we present new theoretical results inspired by the TRPO paper.
\end{comment}
\section{Estimators difference}
\label{appendix-sec:analyzing-the-difference-setimators}
We denote $\pi_\theta$ a parameterized policy with parameter vector $\theta$. We defined $\mathcal{L}$ and $\mathcal{L}^\prime$ as follows:
\begin{align}
 \mathcal{L}_1(\theta, \theta_0, A_{\pithetazero}) &= \EX_{\pithetazero}\left[\sum_{t=0} \gamma^t(\frac{\pitheta(A_t, S_t)}{\pithetazero(A_t, S_t)} - 1) A_{\pithetazero}(S_t, A_t), S_0= s  \right], \\
 \mathcal{L}_2(\theta, \theta_0, A_{\pithetazero}) &= \EX_{\pithetazero}\left[\sum_{t=0} \gamma^t(1 - \frac{\pithetazero(A_t, S_t)}{\pitheta(A_t, S_t)}) \ A_{\pithetazero}(S_t, A_t), S_0= s  \right].   
\end{align}
Differentiating with respect to $\theta$:
\begin{align}
 \nabla_\theta \mathcal{L}_1(\theta, \theta_0, A_{\pithetazero}) &= \EX_{\pithetazero}\left[\sum_{t=0} \gamma^t \nabla \pitheta(A_t, S_t) \frac{1}{\pithetazero(A_t, S_t)}  A_{\pithetazero}(S_t, A_t), S_0= s  \right], \\
 \nabla_\theta \mathcal{L}_2(\theta, \theta_0, A_{\pithetazero}) &= \EX_{\pithetazero}\left[\sum_{t=0} \gamma^t \nabla \pitheta(A_t, S_t) \frac{\pithetazero(A_t, S_t)}{\pitheta(A_t, S_t)^2} \ A_{\pithetazero}(S_t, A_t), S_0= s  \right].
\end{align}
To compare $\mathcal{L}_1$ and $\mathcal{L}_2$, we plot the ratios $r=\frac{1}{\pithetazero}$ and $r^\prime=\frac{\pithetazero}{\pitheta^2}$ with different values of $\pithetazero=0.2, 0.5, 0.8$.
\begin{figure}[H]
  \centering
  \begin{minipage}[b]{0.4\textwidth}
    \includegraphics[width=\textwidth]{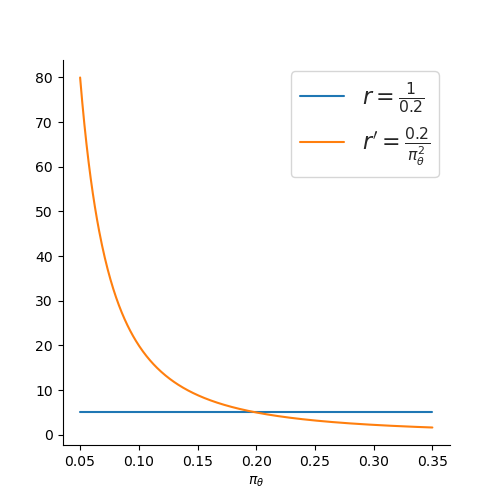}
  \end{minipage}
  \hfill
  \begin{minipage}[b]{0.4\textwidth}
    \includegraphics[width=\textwidth]{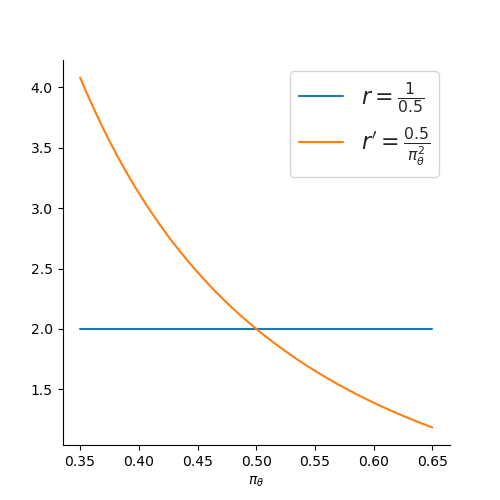}
  \end{minipage}

    \begin{minipage}[b]{0.4\textwidth}
        \includegraphics[width=\textwidth]{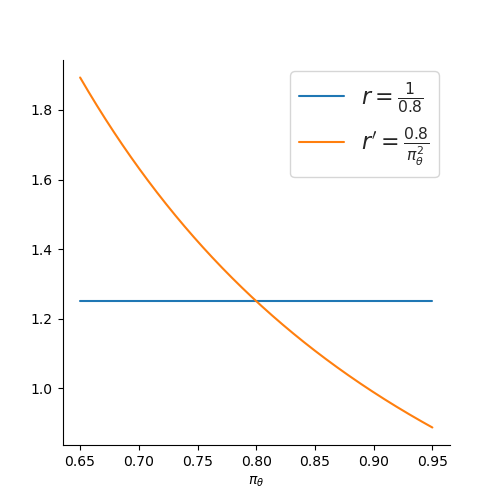}
    \end{minipage}
    \caption{Comparing $\mathcal{L}_1$ and $\mathcal{L}_2$ by plotting $r=\frac{1}{\pithetazero}$ and $r^\prime=\frac{\pithetazero}{\pitheta^2}$ } with different values of $\pithetazero=0.2, 0.5, 0.8$.
\end{figure}

\section{Safe policy iteration}\label{appendix-chapter:SafePolicyIteration}
\subsection{Generalized advantage estimation}
\label{appendix-generalized-advantage-estimation}
We can leverage the findings of the GAE paper to estimate $A^{rc, k}_\pi$ by proving shared properties between $V$ and $V^{rc, k}$. We use the same notation and definitions in \cite{GAE}. Let $r^\prime_t = r_t Q^c_t - \beta c_t$ and $\delta^V_t = r^\prime_t + \gamma V^{rc, k}_{t+1} - V^{rc, k}_t$.\\ 

\begin{alignat*}{2}
    \hat{A}^{(1)}_t &= \delta^V_t &&= -V^{rc, k}_t + r^\prime_t + \gamma V^{rc, k}_{t+1};\\
    \hat{A}^{(2)}_t &= \delta^V_t + \gamma \delta^V_{t+1} &&= -V^{rc, k}_t + r^\prime_t + \gamma  r^\prime_{t+1} + \gamma^2 V^{rc, k}_{t+2};\\
    \hat{A}^{(3)}_t &= \delta^V_t + \gamma \delta^V_{t+1} + \gamma^2 \delta^V_{t+2} &&= -V^{rc, k}_t + r^\prime_t + \gamma  r^\prime_{t+1} + \gamma^2  r^\prime_{t+2} + \gamma^{t+3} V^{rc, k}_{t+3};\\
    & \cdots\\
    \hat{A}^{(q)}_t & \coloneqq \sum_{l=0}^{q-1} \gamma^l \delta^V_{t+l} = -V^{rc, k}_t +r^\prime_{t} + && \gamma r^\prime_{t+1} + \gamma^2 r^\prime_{t+1} + ... + \gamma^{q-1} r^\prime_{t+q-1} + \gamma^q  V^{rc, k}_{t+q}. 
\end{alignat*}
Therefore, the following equation can be used to estimate $A^{rc, k}_\pi$:\\
\begin{equation}
    \hat{A}^{\text{GAE}(\gamma, \lambda)} = \sum_{l=0} (\gamma \lambda)^l \delta^V_{t+l}.
\end{equation}

\subsection{Practical implementation}\label{appendix-chapter:PraticalImplementation}
% \subsection{Safety critic policy optimization}
After generating trajectories using $\pi$, we can compute $\hat{A}^{r, r}_\pi(s, a)$ and $\hat{A}^c_\pi(s, a)$ for every state-action pair. We denote a parameterized policy by $\pi_{\theta}$. We refer to the The initial policy by $\pi_{\theta_0}$.\\
For notational convenience, we define $r_t(\theta) = \frac{\pi_\theta(A_t \kgiven S_t)}{\pi_{\theta_0}(A_t\kgiven S_t)} - 1$, $r_t^\prime(\theta) = 1-\frac{\pi_{\theta_0}(A_t \kgiven S_t)}{\pi_{\theta}(A_t\kgiven S_t)}$ and $A_t = A(S_t, A_t)$. We follow the same clipping strategy in \cite{ppo}. We define clipped version of $\mathcal{L}_1$ and $\mathcal{L}_2$:
\begin{align*}
     \mathcal{L}_1^\text{CLIP}(\theta, \theta_0, A_{\pithetazero}) &= \EX_{\pithetazero}\left[\sum_{t=0} \min(r_t(\theta) A_t, \text{clip}(r_t(\theta), -\epsilon, +\epsilon) A_t )\kgiven S_0= s  \right],\\
    \mathcal{L}_2^\text{CLIP}(\theta, \theta_0, A_{\pithetazero}) &= \EX_{\pithetazero}\left[\sum_{t=0} \min(r_t^\prime(\theta) A_t, \text{clip}(r_t^\prime(\theta), -\epsilon, +\epsilon) A_t )\kgiven S_0= s  \right],
\end{align*}
where $\epsilon$ is hyperparameter, usually chosen $\epsilon=0.2$. 

To approximate $\pi$, $V^c$, and $V^{rc, k}$, we use three fully-connected MLPs with two hidden layers and tanh nonlinearities. The policy network outputs the mean of a Gaussian distribution with variable standard deviations for continuous tasks, as described in \cite{trpo,dua}.
When the reward is negative, a reward bias is added such that for all state $s$ and action $a$, $r^\prime(s, a) = r(s, a) + b > 0$.\\
The output of the safety critic MLP is tanh clipped to the range $[0, 1]$. This ensures the safety critic is pessimistic when randomly initialized: $V^c_{\pi_0}(s) \approx 0$.
In the following algorithm, we refer to $\mathcal{L} = \mathcal{L}_1$ or $\mathcal{L} = \mathcal{L}_2$. We start by a randomly initialized policy $\pi_{\theta}$
\begin{algorithm}[H]
\caption{Safety critic policy optimization (SCPO)} 
\label{appendix-alg:SAFEPPO}
\begin{algorithmic}[1]
\FOR{iteration$=0, 1, 2, 3...$}
    \STATE Sample $(a_t, s_t) \sim \pi_{\theta_0}$ for T timesteps.
    \STATE Evaluate $\hat{Q}^c(s_t, a_t)$.
    \STATE $r(s, a) \leftarrow r(s,a) + b$ (positive reward, $b \geq 0$).
    \STATE $r(s, a) \leftarrow r(s,a) \ \hat{Q}^c(s, a)^k - \beta (1-Q^c(s, a)^k) c(s,a)$.
    \STATE Use GAE to estimate $\hat{A}^{rc, k}_{\pithetazero}(s_t, a_t)$.
    \FOR{epoch$=0, 1, 2, 3...$}
        \STATE  Optimize $\mathcal{L}^\text{CLIP}(\theta, \theta_0, A^{rc, k}_{\pithetazero})$ wrt $\theta$.
        \STATE Critic update: $V^{rc, k}$ and $V^c$.
    \ENDFOR
    \STATE $\theta_0 \leftarrow \theta$.
\ENDFOR
\end{algorithmic}
\end{algorithm}
When $k=0$ and $\beta > 0$, then $\beta$ is a Lagrange multiplier for the constrained safe RL problem:\\
\begin{equation*}
    \max_{\pi} V_\pi(s_0), \qquad \text{s.t} \quad \EX_\pi\left[\sum_{t=0} c(S_t) \kgiven S_0= s_0\right] \leq c_0.
\end{equation*}

\section{Experiments}\label{chapter:Experiments}
In this section, we evaluate the performance of our algorithm (SCPO) by comparing it to TRPO-L \cite{Ray}, CPO \cite{cpo}, PDO, and PCPO \cite{PCPO}. We use the ball agent from safety bullet gym \cite{BulletSafetyGym} on four tasks: circle, reach, gather, run. The experiment results demonstrate the effectiveness of our approach. We also highlight the importance of augmenting the state representation using the cumulative cost as described in Section \ref{section:augment_state}.
\subsection{Environments} \label{appendix-section:environments-settings}
\begin{figure}[!ht]
  \centering
  \begin{minipage}[b]{0.41\textwidth}
    \includegraphics[width=\textwidth]{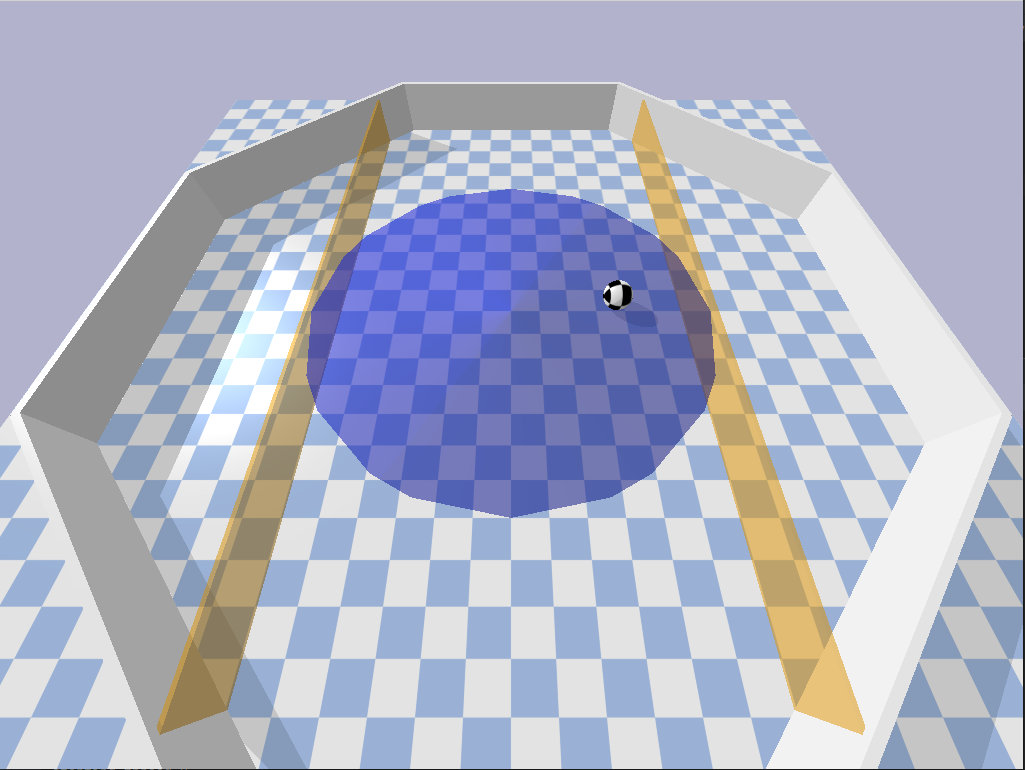}
    \caption{SafetyBallCircle}.
    \label{env:SafetyBallCircle}
  \end{minipage}
  \hfill
  \begin{minipage}[b]{0.41\textwidth}
    \includegraphics[width=\textwidth]{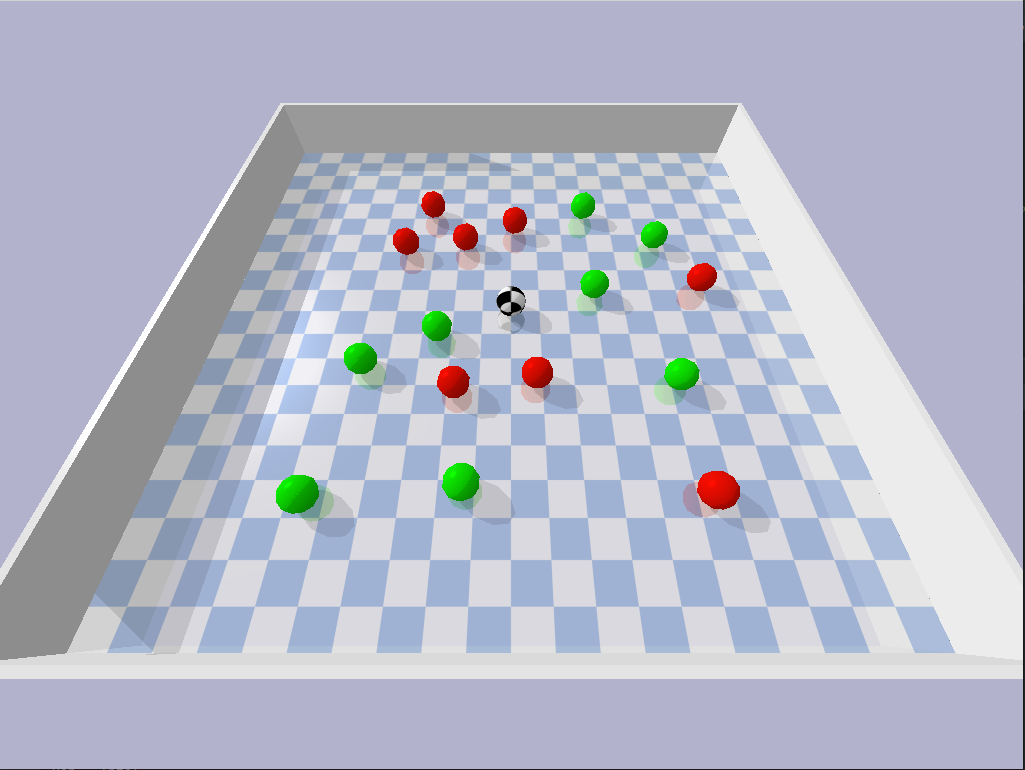}
    \caption{SafetyBallGather}.
    \label{env:SafetyBallGather}
  \end{minipage}

    \begin{minipage}[b]{0.41\textwidth}
        \includegraphics[width=\textwidth]{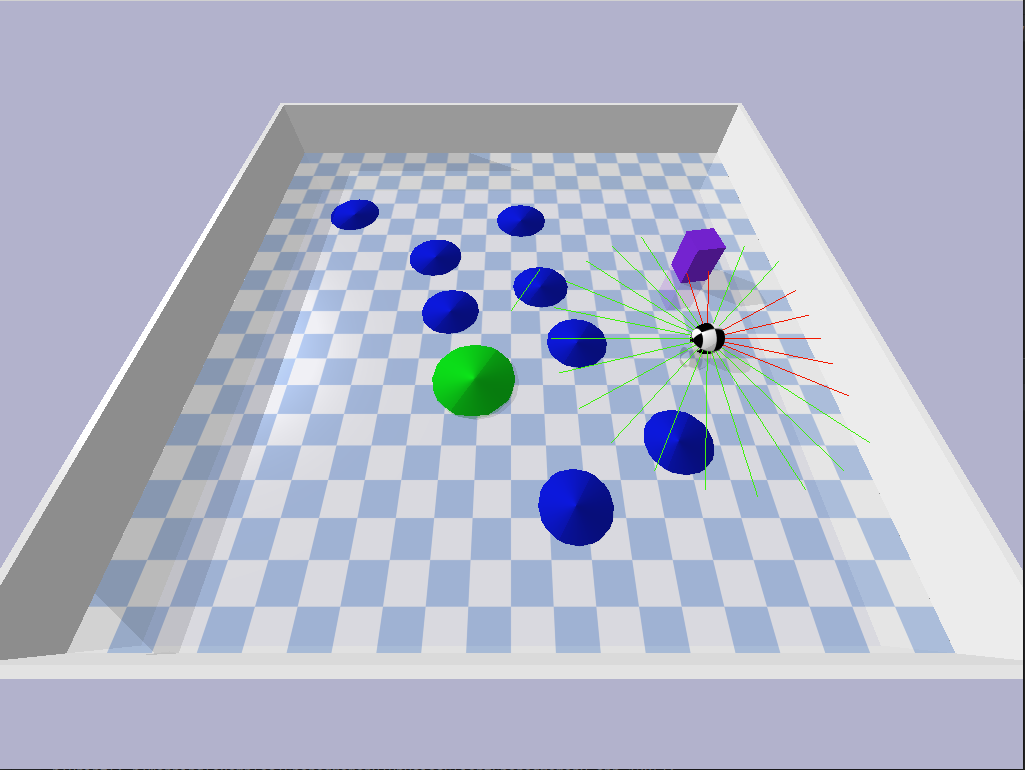}
        \caption{SafetyBallReach}.
        \label{env:SafetyBallReach}

    \end{minipage}
      \hfill
    \begin{minipage}[b]{0.41\textwidth}
        \includegraphics[width=\textwidth]{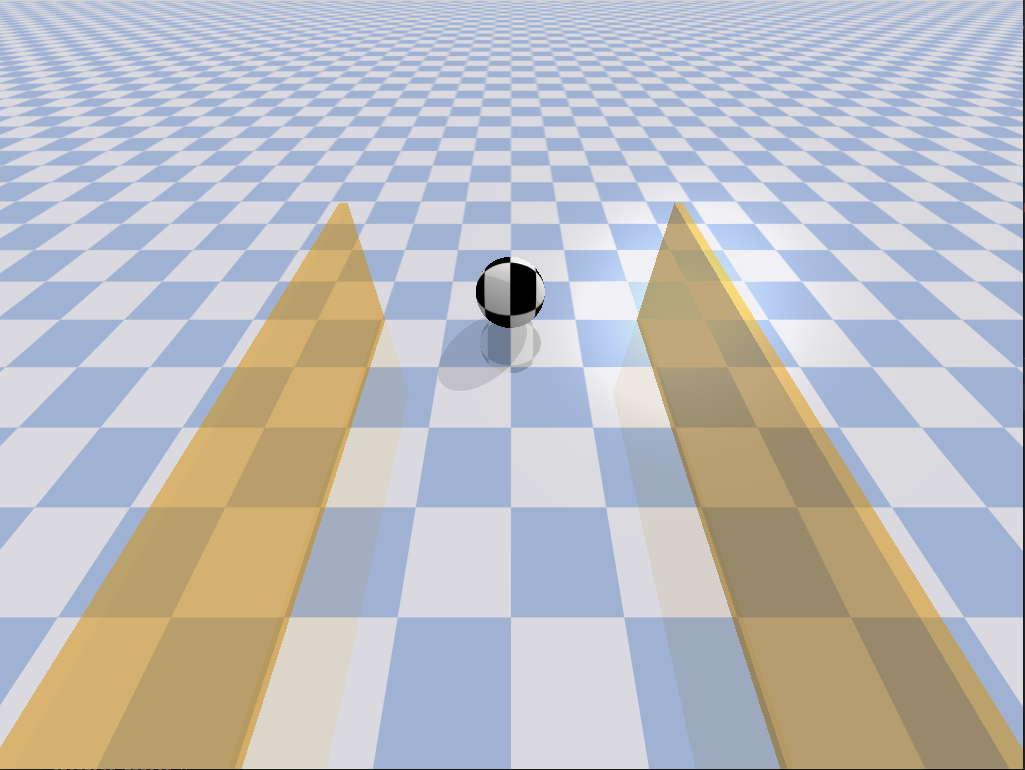}
        \caption{SafetyBallRun}.
        \label{env:SafetyBallRun}
    
    \end{minipage}
    \begin{minipage}[b]{0.41\textwidth}
        \includegraphics[width=\textwidth]{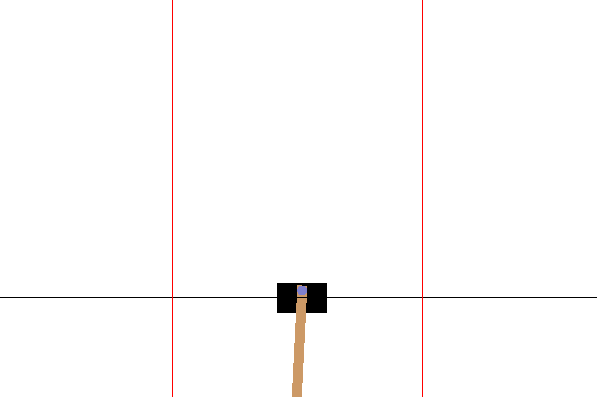}
        \caption{CartSafe}.
        \label{env:CartSafe}
        
    \end{minipage}
\end{figure}

We use a subset of environments from a free and open-source framework called Bullet-Safety-Gym \cite{BulletSafetyGym}.\\
The agent is a ball which can move freely on the xy-plane. The shape of the observations space is $\mathbb{R}^9$, which contains the position $x\in \mathbb{R}^3$ and the velocity $\dot{x}\in \mathbb{R}^3$.Actions are applied as forced $a\in [-1, 1]^2$.
\\\\
\textbf{\hyperref[env:SafetyBallCircle]{SafetyBallCircle}}: The goal is to move clockwise without leaving the safe area, as shown in Figure~\ref{env:SafetyBallCircle}.  \\
Reward: The reward is maximized when the agent moves clockwise as fast as possible. $r(s) = \frac{v^T[-y, x]}{1+3|r_\text{agent} - r_\text{circle}|}$.\\
Cost: A cost of 1 is incurred when the agent is outside the bounds denoted by two vertical lines, e.g. $c(s) = \mathbf{1}[|x| \geq x_\text{lim}]$.\\\\
\textbf{\hyperref[env:SafetyBallGather]{SafetyBallGather}}: The Agent is spawned randomly and incentivized to collect blue balls and avoid collecting red ones, as shown in Figure~\ref{env:SafetyBallGather}.\\
Reward: The agent receives a reward of 10 when it comes in contact with a blue ball and collects it.\\
Cost: The agent receives immediate cost 1 when it comes in contact with a red ball, collecting it.\\\\
\textbf{\hyperref[env:SafetyBallRun]{SafetyBallRun}}: The agent is incentivized to run as fast as possible in the x direction, as shown in Figure~\ref{env:SafetyBallRun}.\\
Reward: Increases proportionally to the velocity in the x direction.\\
Cost: Received when the agent exceeds a velocity threshold of 2.5 or when leaving the non-physical boundary.\\\\
\textbf{\hyperref[env:SafetyBallReach]{SafetyBallReach}}: The agent is incentivized to chase the area marked in green while avoiding the rectangular obstacle and areas marked in blue, as shown in Figure~\ref{env:SafetyBallReach}.\\
Reward: Shaped reward based on the Euclidean distance between the agent and the goal. Sparse when the agent comes in contact with the green area.\\
Cost: sparse cost received when the agent comes in contact with the rectangular obstacle or walks over the blue zone.\\\\
\textbf{\hyperref[env:CartSafe]{CartSafe}}: The agent is a cart. Actions are discrete $a\in \{0, 1\}$. The agent is incentivized to balance the pole upright while staying within bounds, as shown in Figure~\ref{env:CartSafe}.\\
Reward: Shaped reward that scales with the upright pole position: $r(s) = 1 + cos(Pole angle)$.\\
Cost: $c(s) = \mathbf{1}[-1 < \textbf{cart position} < 1]$.\\
The episode terminates if the distance between the cart and the centre exceeds 2.4.

\clearpage

\subsection{Bullet-Safety-Gym benchmark}
\subsubsection{Other safe reinforcement learning algorithms}
\label{appendix-safe-baselines-introduction}
We compare our algorithm \hyperref[appendix-alg:SAFEPPO]{SCPO} to the following on-policy algorithms:
\begin{itemize}
 \item Trust-Region Policy Optimization (TRPO) \cite{trpo}: unconstrained policy optimization algorithm. A line search is used to determine policy update step size.
\item TRPO-L: Uses the TRPO objective and transforms the constrained optimization problem to an unconstrained one using Lagrange multiplier. The Lagrange multiplier is learnable and changes during policy iteration.
\item Constrained Policy Optimization (CPO) \cite{cpo}: Computes Lagrange multiplier for each policy iteration step. Uses the trust region objective.
\item Primal-dual Optimization (PDO): Uses a Lagrange multiplier that can be learned and retains its state.
\item Projection-based Constrained Policy Optimization (PCPO) \cite{PCPO}: is a two-stage optimization technique based on CPO. The first step updates the parameters without constraints. The second addresses constrained violation by projecting the policy parameters on the constraint set.
\end{itemize}
We use TRPO to estimate an upper bound of the return and cost when safety is not considered. The above algorithms and their benchmarks are provided by Sven Gronauer and described in more detail in his technical report  \cite{BulletSafetyGym}.\\
A discount factor $\gamma = 0.99$ was used, and $B=32000$ environment steps were collected. To train safe SCPO, we use $B=32768$ environment steps. Therefore, we scale our plots to match the other algorithms and omit the x-axis label.\\
The neural network architecture of all algorithms consists of a multi-layer perceptron (MLP) with two hidden layers of size 64, followed by a tanh non-linearity. The Adam optimizer \cite{adam} is used in our implementation.\\
Each algorithm was evaluated on four different random seeds. If an algorithm violates the safety constraint over all hyperparameters, we pick the hyperparameter with the least average cumulative cost over the last 50 iterations. We use Stable Baselines 3 \cite{sb3} as the foundation for our implementation.

\subsubsection{CartSafe}
\label{subsection:CartSafe}
The action space of CartSafe is discrete. Unlike the other environments, a randomly initialized policy violates the safety constraint with high probability, e.g. $Q_{\pi_0}^c(s, a) \approx 0$ for all state $s$ and action $a$. Under this condition, using the objective function $V^{r,k}$ might lead to difficulty in learning a safe behavior. However, this problem is solved by introducing the objective function $V^{rc, k}$ as described in Section \ref{subsec:Canceling-Unsafe-Reward}.
% \ref{section:reducing-cost}.

\begin{figure}[H]
  \centering
  \begin{minipage}[]{0.495\textwidth}
    \includegraphics[width=\textwidth]{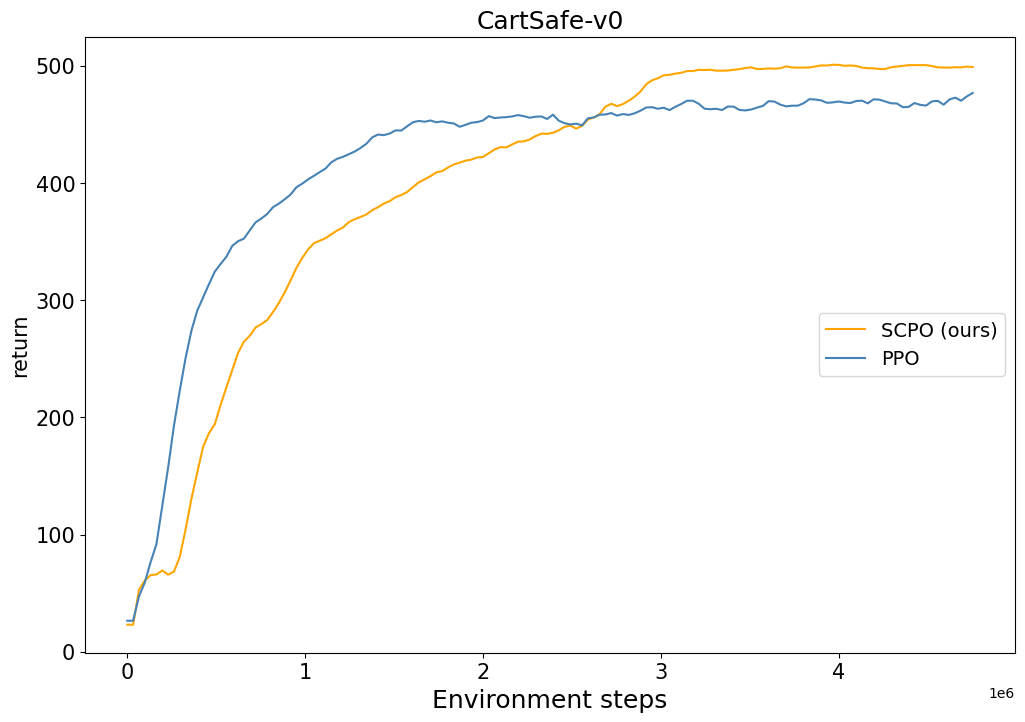}
  \end{minipage}
  \hfill
  \begin{minipage}[]{0.495\textwidth}
    \includegraphics[width=\textwidth]{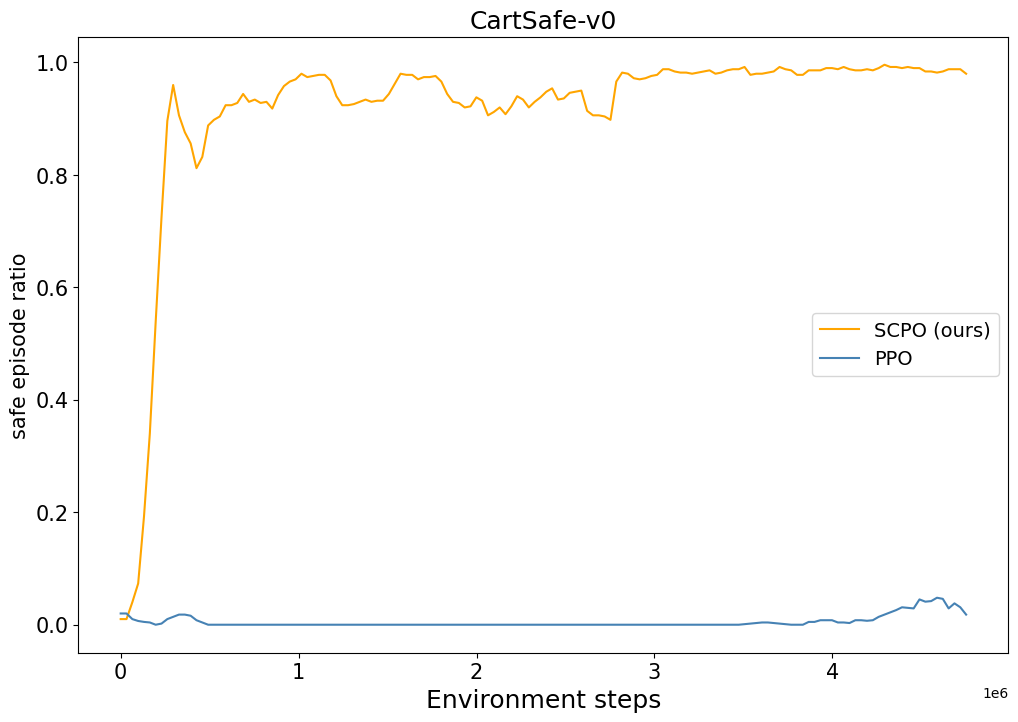}
  \end{minipage}

    \begin{minipage}[]{0.495\textwidth}
    \includegraphics[width=\textwidth]{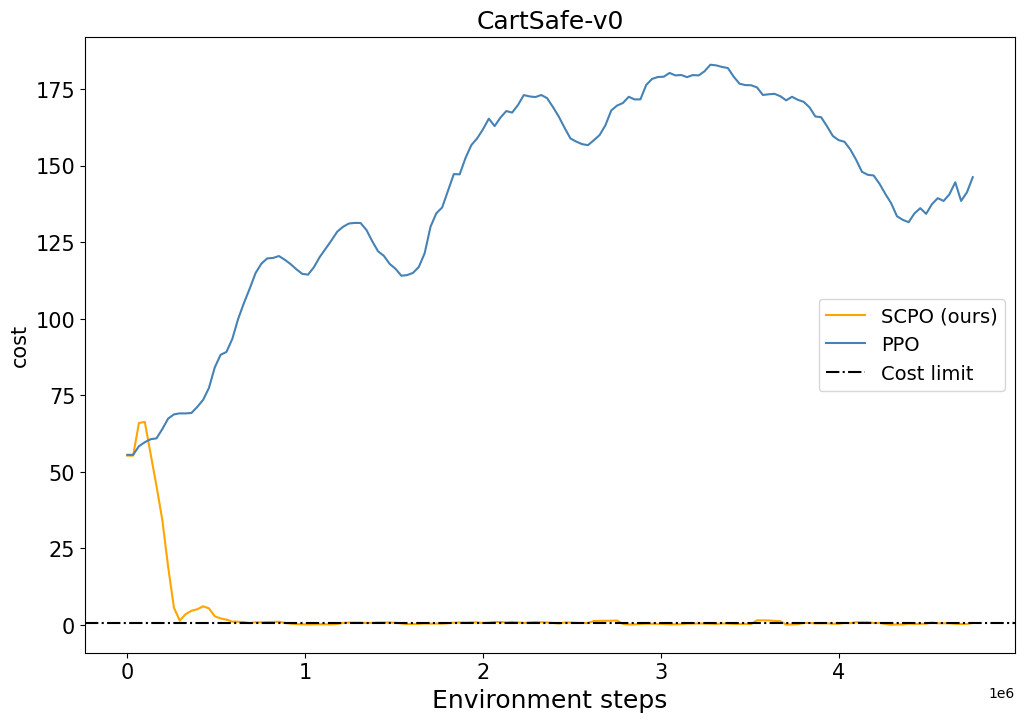}
  \end{minipage}
  \hfill
  \begin{minipage}[]{0.495\textwidth}
    \includegraphics[width=\textwidth]{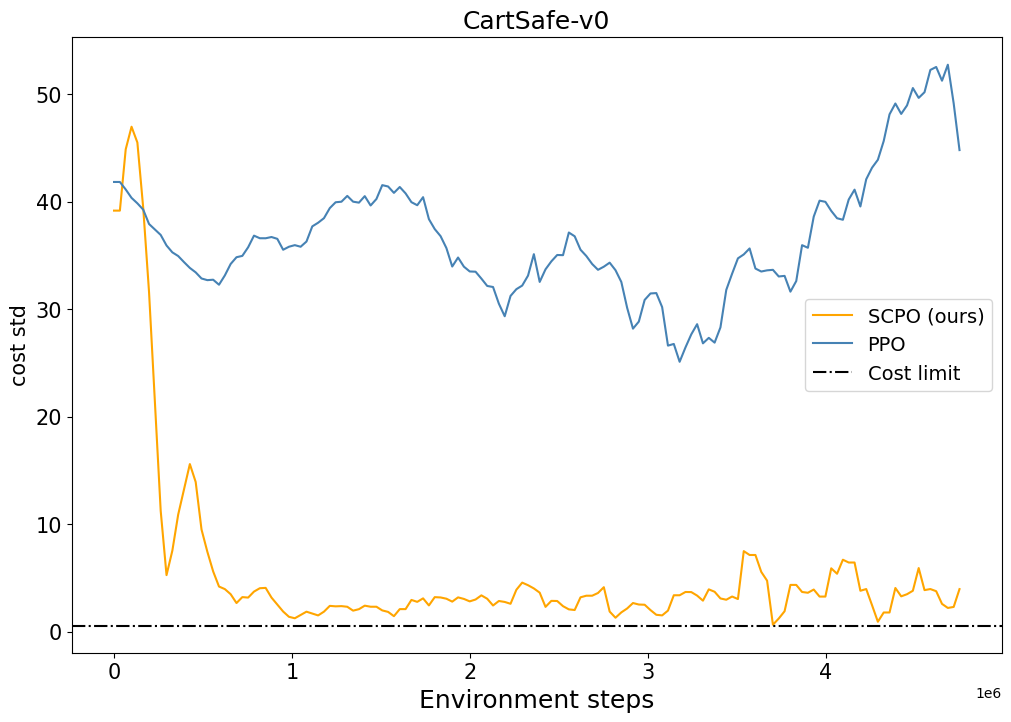}
  \end{minipage}
    \caption{The learning curve of SCPO and PPO using the environment CartSafe-v0.}
    \label{figure:benchmark-cart-safe}
\end{figure}

When all state-action pairs are unsafe, using $V^{rc, k}$ is equivalent to minimizing the cumulative cost, eventually leading to a safe policy.\\
Figure \ref{figure:benchmark-cart-safe} demonstrates the ability of SCPO to quickly find a safe policy and simultaneously improve the return. Safety critic plays a crucial role, enabling a seamless transition from cost reduction to return maximization. 

% \begin{comment}
\subsection{Augmenting state effect}
\label{subsection:augmenting-state-effect}
We investigate the effect of state augmentation by training a policy using \hyperref[alg:SAFEPPO]{safe ppo} on the environment 
\textbf{\hyperref[env:SafetyBallRun]{SafetyBallRun}}. The agent receives a reward proportional to its velocity. A cost of $1$ is incurred when the agent's velocity exceeds $2.5$. We train two policies: one using augmented states (as explained in Section \ref{section:augment_state}) and another using normal states. Both policies share the same hyperparameters.\\
\begin{figure}[!ht]
  \centering
  \begin{minipage}[]{0.495\textwidth}
    \includegraphics[width=\textwidth]{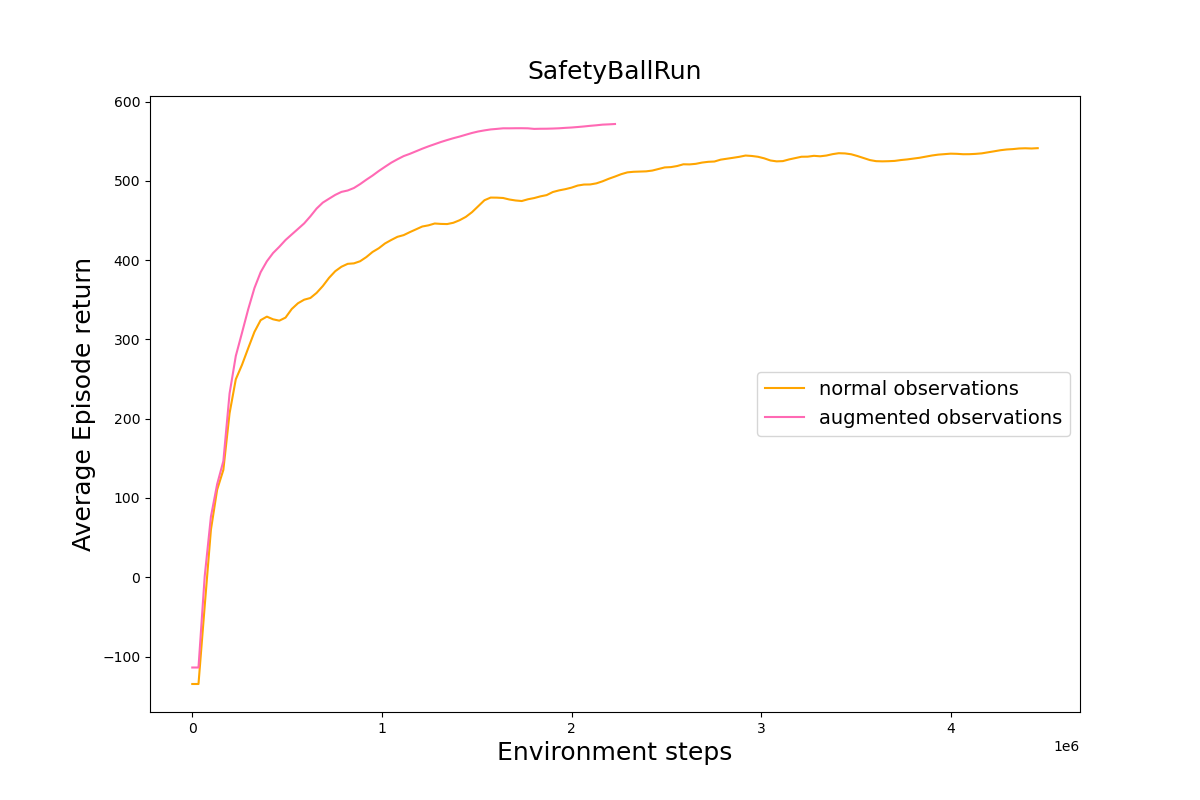}
  \end{minipage}
  \hfill
  \begin{minipage}[]{0.495\textwidth}
    \includegraphics[width=\textwidth]{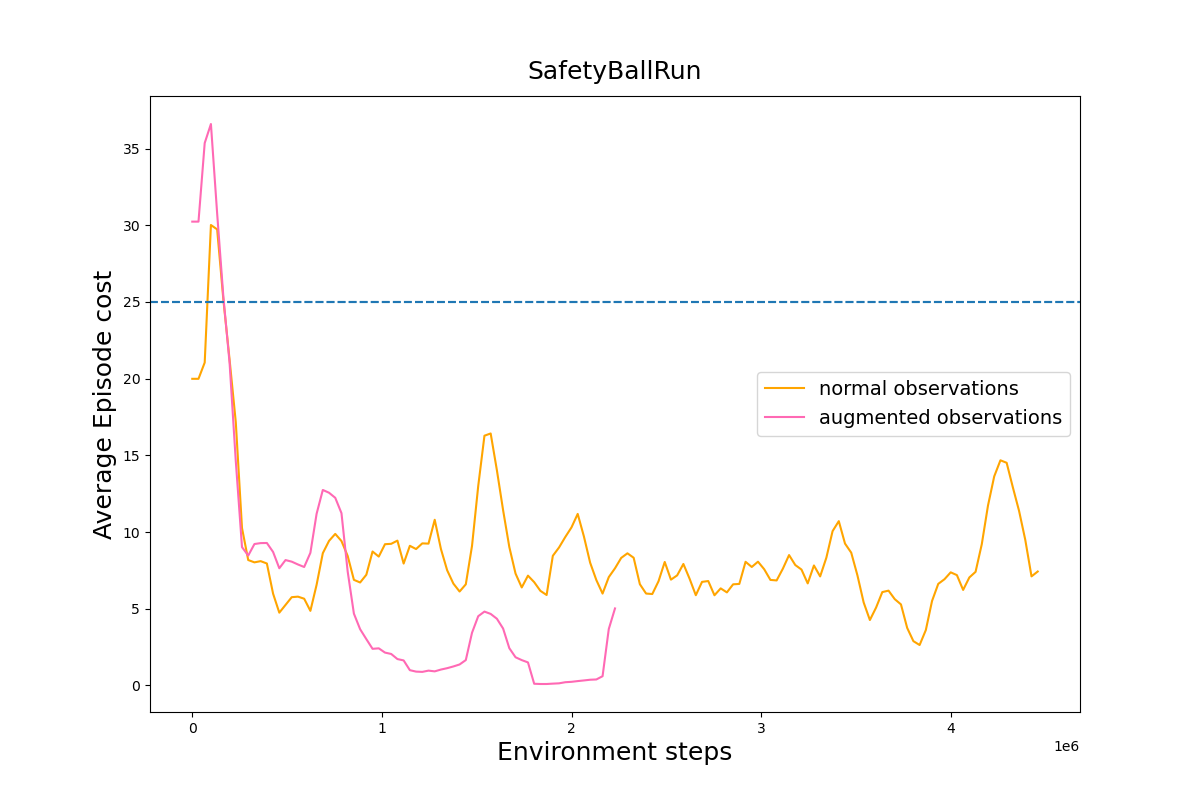}
  \end{minipage}
  \caption{SafetyBallCircle return and cumulative cost.}
\label{fig:state_augment_ball_run}
\end{figure}
\\
The policy trained with augmented states converges faster to the maximum return and requires training samples, as shown in figure \ref{fig:state_augment_ball_run}. While both policies achieve a similar return, they exhibit different behavior. 
\begin{figure}[!ht]
  \centering
  \begin{minipage}[]{0.495\textwidth}
    \includegraphics[width=\textwidth]{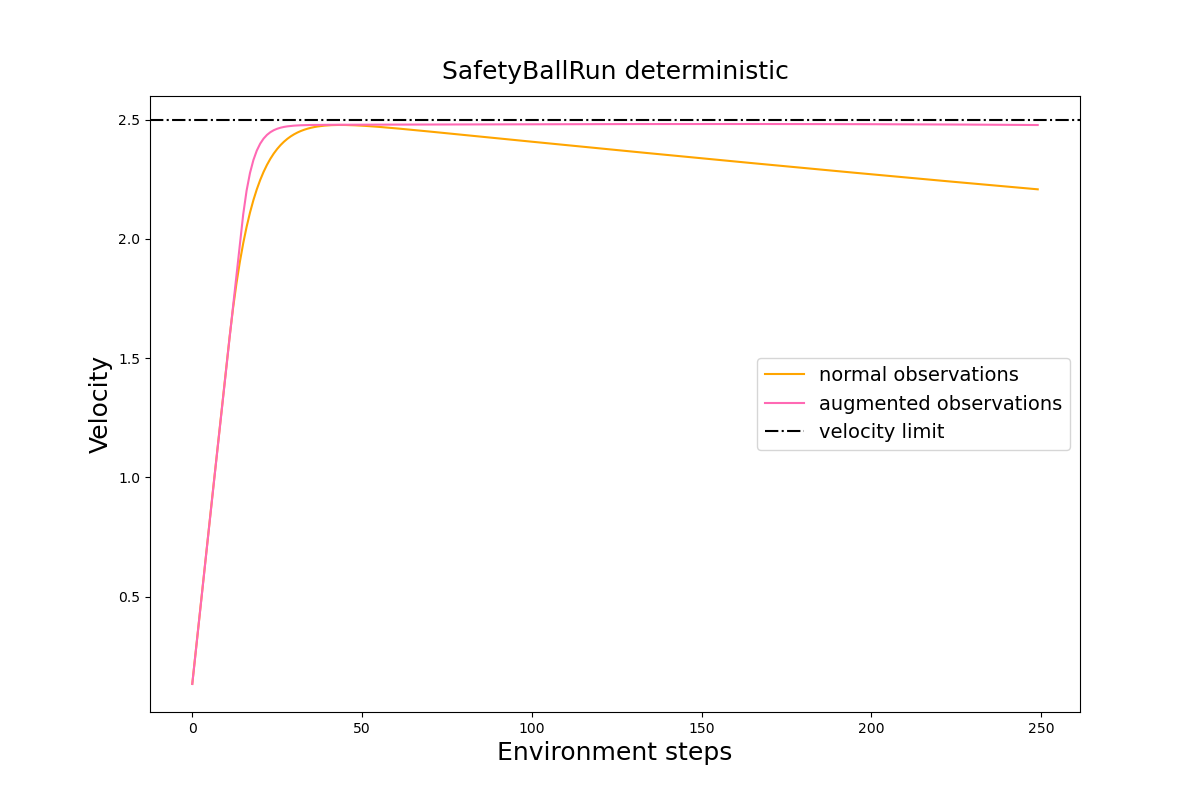}
  \end{minipage}
  \hfill
  \begin{minipage}[]{0.495\textwidth}
    \includegraphics[width=\textwidth]{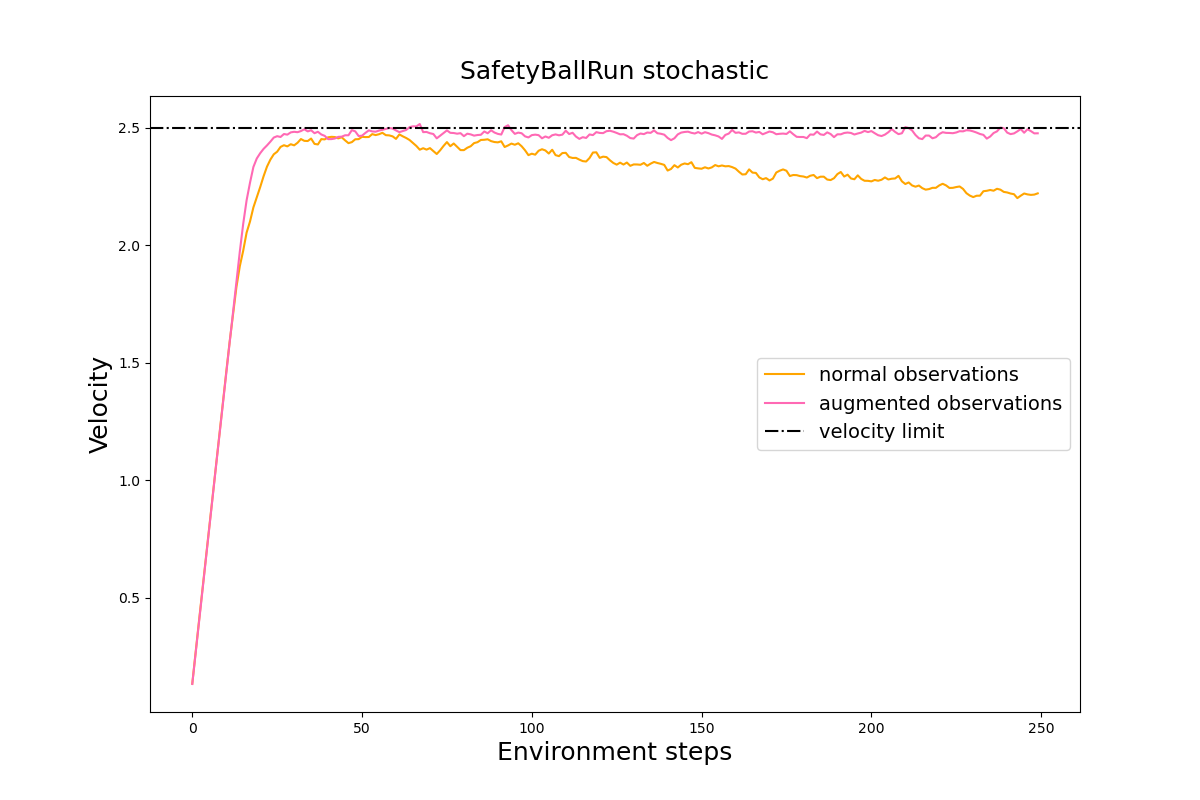}
  \end{minipage}
  \caption{SafetyBallCircle velocity.}
\label{fig:state_augment_ball_run}
\end{figure}
\\
The policy trained with augmented states maintains constant velocity after accelerating. It knows the current cumulative cost. Exceeding the velocity limit does not immediately violate the safety constraint. The same unaugmented state can cause constraint violation or be safe depending on the current cumulative cost value. Therefore, using augmented states removes this ambiguity. The policy trained with normal states accelerates then decelerates. The policy likely anticipates the constraint violation but does not have the means to correctly identify risky states.
% \end{comment}

 \section{Details of Implementing Experiments}
 The experiment parameters are provided in Tables \ref{demo-table:safety-bound} and \ref{demo-table:scpo-hyparameters}. A server with one AMD CPU (eight-core processor) and one NVIDIA GTX 2060 GPU is used to run the experiments.
\begin{table}[!h]
\begin{center}
\begin{tabular}{  c  c  c  } 
  \hline
  Environment Name & Cost limit & Maximum Episode Length \\
\hline
  SafetyBallCircle-v0   & 25 & 250 \\ 
  SafetyBallRun-v0  & 25 & 250 \\ 
  SafetyBallGather-v0 & 0.2 & 250 \\
  SafetyBallReach-v0 & 10 & 250 \\
  CartSafe-v0 & 1 & $[0, 300]$ \\
  \hline
\end{tabular}
\vspace{5pt}
\caption{The cost limit and episode length used in the environments.}
\label{demo-table:safety-bound}
\end{center}
\end{table}

% \begin{tabular}{ | m{12 em} | m{4em} | m{5em} | m{4em} | m{4em} | m{4em} |} 

\begin{table}[!h]
% \begin{center}
\scriptsize{
\begin{adjustbox}{width=0.9\textwidth,center}
\begin{tabular}{  c  c  c  c  c  c } 
  \hline
    Hyper-parameter & BallCircle & BallGather & BallRun & BallReach & CartSafe \\
  \hline
    Batch-size & 64 & 64 & 64 & 64 & 64\\
    Epochs & 5 & 5 & 5 & 5 & 5\\
    Learning rate & 2e-4 & 2e-4 & 2e-4 & 2e-4 & 2e-4\\
    Optimizer & Adam & Adam & Adam & Adam & Adam\\
    Timesteps T & 32768 & 32768 & 32768 & 32768 & 32768\\
    Entropy co-efficient & 0.01 & 0.01 & 0.005& 0.01 & 0.001\\
    Clip range & 0.2 & 0.2 & 0.2 & 0.2 & 0.2\\
    GAE factor rewards $\lambda$ & 0.95 & 0.95 & 0.95 & 0.95  & 0.95\\
    Discount $\gamma$ &  0.99 & 0.99 & 0.99 & 0.99 & 0.99\\
    Safety discount $\gamma$ (\ref{appendix-section:V-C-ESTIMATE}) & 0.995 & 0.995 &  0.995 &  0.995 &  0.995\\
    Reward bias $b$ & 1.5 &  0.05 & 1 & 0.1 & 0\\
    $k$ ($V^{rc, \textbf{$k$}}$) & 2 & 4 & 4 & 4 & 5\\
    Cost factor $\beta$ & 0 & 15 & 0.5 & 0 & 3\\ 
\hline
% \vspace{0.1pt}
\end{tabular}
\end{adjustbox}
}
\vspace{5pt}
\caption{Hyper-parameters used to train SCPO.}
\label{demo-table:scpo-hyparameters}
% \end{center}
\end{table}
\clearpage

\end{document}